\def\year{2022}\relax

\documentclass[letterpaper]{article}  % DO NOT CHANGE THIS
\usepackage{aaai22}                   % DO NOT CHANGE THIS
\usepackage{times}                    % DO NOT CHANGE THIS
\usepackage{helvet}                   % DO NOT CHANGE THIS
\usepackage{courier}                  % DO NOT CHANGE THIS
\usepackage[hyphens]{url}             % DO NOT CHANGE THIS
\usepackage{graphicx}                 % DO NOT CHANGE THIS
\usepackage{natbib}                   % DO NOT CHANGE THIS AND DO NOT ADD ANY OPTIONS TO IT
\usepackage{caption}                  % DO NOT CHANGE THIS AND DO NOT ADD ANY OPTIONS TO IT
\DeclareCaptionStyle{ruled}{labelfont=normalfont,labelsep=colon,strut=off} % DO NOT CHANGE THIS
\usepackage{array}     % table alignment
\usepackage{amsmath}   % math symbols and environments
\usepackage{booktabs}  % professional-quality tables
\usepackage{graphbox}  % graphics alignment
\usepackage{ifthen}    % ifthenelse macro
\usepackage{lineno}    % line numbers for submission
\usepackage{ltxcmds}   % LaTeX utility macros (e.g., \ltx@ifpackageloaded)
\usepackage{multirow}  % multi-row cells in tables
\usepackage{tikz}      % shape drawing
\usepackage{xcolor}    % RGB color definitions

\newcommand\type{preprint}

\newcommand{\ifsubmission}[2]{\ifthenelse{\equal{\type}{submission}}{#1}{#2}}
\newcommand{\iffinal}[2]{\ifthenelse{\equal{\type}{final}}{#1}{#2}}
\newcommand{\ifpreprint}[2]{\ifthenelse{\equal{\type}{preprint}}{#1}{#2}}

\ifsubmission{}{\iffinal{}{\ifpreprint{}{\PackageError{}{Unknown type}{}}}}

\ifpreprint{\usepackage[backref=page]{hyperref}}{}

\ifsubmission{\linenumbers}{}

\nocopyright

\makeatletter\newcommand{\IfPackageLoaded}[3]{\ltx@ifpackageloaded{#1}{#2}{#3}}\makeatother

\newcommand\mailto[1]{\IfPackageLoaded{hyperref}{\href{mailto:#1}{#1}}{#1}}
\newcommand{\meansd}[2]{${#1}\pm#2$}

\definecolor{legend_blue}{RGB}{31,119,180}
\definecolor{legend_orange}{RGB}{255,127,14}
\DeclareRobustCommand{\square}[2][0ex]{
  \raisebox{#1}{\raisebox{0.1465ex}{\tikz\draw[#2,fill=#2] (0,0) rectangle (0.707ex, 0.707ex);}}}
\DeclareRobustCommand{\diamond}[2][0ex]{
  \raisebox{#1}{\tikz\draw[#2,fill=#2,rotate=45] (0,0) rectangle (0.707ex, 0.707ex);}}

\DeclareMathOperator{\sign}{sign}
\DeclareMathOperator*{\argmax}{argmax}
\DeclareMathOperator*{\argmin}{argmin}

\IfPackageLoaded{hyperref}{
  \hypersetup{
    pdfinfo={
      Title={Measuring the Contribution of Multiple Model Representations in Detecting Adversarial Instances},
      TemplateVersion={2022.1}
    }
  }
  \ifsubmission{\hypersetup{pdfinfo={Author={Anonymous Author(s)}}}}
    {\hypersetup{pdfinfo={Author={Daniel Steinberg, Paul Munro}}}}
}{
  \pdfinfo{
    /Title (Measuring the Contribution of Multiple Model Representations in Detecting Adversarial Instances)
    /TemplateVersion (2022.1)
  }
  \ifsubmission{\pdfinfo{/Author (Anonymous Author(s))}}
    {\pdfinfo{/Author (Daniel Steinberg, Paul Munro)}}
}

\title{
  Measuring the Contribution of Multiple Model \\
  Representations in Detecting Adversarial Instances
}

\ifsubmission{\author{Anonymous Author(s)}}{
  \author{
    Daniel Steinberg,\!\textsuperscript{\rm 1}
    Paul Munro\textsuperscript{\rm 2}
  }
}

\ifsubmission{\affiliations{Affiliation \\ Address \\ email}}{
  \affiliations{
    \textsuperscript{\rm 1} Intelligent Systems Program, University of Pittsburgh \\
    \textsuperscript{\rm 2} School of Computing and Information, University of Pittsburgh \\
    {\mailto{das178@pitt.edu}}, {\mailto{pwm@pitt.edu}}
  }
}

\begin{document}

\maketitle

\begin{abstract}

% Add Abstract link to Contents.
\addcontentsline{toc}{section}{Abstract}

Deep learning models have been used for a wide variety of tasks. They are prevalent in computer
vision, natural language processing, speech recognition, and other areas. While these models have
worked well under many scenarios, it has been shown that they are vulnerable to adversarial attacks.
This has led to a proliferation of research into ways that such attacks could be identified and/or
defended against. Our goal is to explore the contribution that can be attributed to using multiple
underlying models for the purpose of adversarial instance detection. Our paper describes two
approaches that incorporate representations from multiple models for detecting adversarial examples.
We devise controlled experiments for measuring the detection impact of incrementally utilizing
additional models. For many of the scenarios we consider, the results show that performance
increases with the number of underlying models used for extracting representations.

Code is available at~\ifsubmission{\url{https://anonymized/for/submission}}%
{\url{https://github.com/dstein64/multi-adv-detect}}.

\end{abstract}

\section{Introduction}
\label{sec:introduction}

Research on neural networks has progressed for many decades, from early work modeling neural
activity~\cite{mcculloch_logical_1943} to the more recent rise of deep
learning~\cite{bengio_deep_2021}. Notable applications include image
classification~\cite{krizhevsky_imagenet_2012}, image generation~\cite{goodfellow_generative_2014},
image translation~\cite{isola_image--image_2017}, and many others~\cite{dargan_survey_2020}. Along
with the demonstrated success it has also been shown that carefully crafted adversarial
instances---which appear as normal images to humans---can be used to deceive deep learning
models~\cite{szegedy_intriguing_2014}, resulting in incorrect output. The discovery of adversarial
instances has led to a broad range of related research including 1)~the development of new attacks,
2)~the characterization of attack properties, and 3)~defense techniques.
\citeauthor{akhtar_threat_2018} present a comprehensive survey on the threat of adversarial attacks
to deep learning systems used for computer vision.

Two general approaches---discussed further in Section~\ref{sec:related_work}---that have been
proposed for defending against adversarial attacks include 1)~the usage of model ensembling and
2)~the incorporation of hidden layer representations as discriminative features for identifying
perturbed data. Building on these ideas, we explore the performance implications that can be
attributed to using representations from multiple models for the purpose of adversarial instance
detection.

\paragraph{Our Contribution} In Section~\ref{sec:method} we present two approaches that use neural
network representations as features for an adversarial detector. For each technique we devise a
treatment and control variant in order to measure the impact of using multiple networks for
extracting representations. Our controlled experiments in Section~\ref{sec:experiments} measure the
effect of using multiple models. For many of the scenarios we consider, detection performance
increased as a function of the underlying model count.

\section{Preliminaries}
\label{sec:preliminaries}

Our research incorporates $l$-layer feedforward neural networks, functions \mbox{$h: \mathcal{X}
  \rightarrow \mathcal{Y}$} that map input $x \in \mathcal{X}$ to output $\hat{y} \in \mathcal{Y}$
through linear preactivation functions $f_i$ and nonlinear activation functions $\phi_i$.
\[
\hat{y} = h(x) = \phi_l \circ f_l \circ \phi_{l-1} \circ f_{l-1} \circ \ldots
\circ \phi_1 \circ f_1(x)
\]

The models we consider are classifiers, where the outputs are discrete labels. For input $x$ and its
true class label $y$, let $J(x, y)$ denote the corresponding loss of a trained neural network. Our
notation omits the dependence on model parameters $\theta$, for convenience.

\subsection{Adversarial Attacks}

Consider input $x$ that is correctly classified by neural network $h$. For an untargeted adversarial
attack, the adversary tries to devise a small additive perturbation $\Delta x$ such that adversarial
input $x^{adv} = x + \Delta x$ changes the classifier's output (i.e., $h(x) \neq h(x^{adv})$). For a
targeted attack, a desired value for $h(x^{adv})$ is an added objective. In both cases, the $L_p$
norm of $\Delta x$ is typically constrained to be less than some threshold~$\epsilon$. Different
threat models---white-box, grey-box, and black-box---correspond to varying levels of knowledge that
the adversary has about the model being used, its parameters, and its possible defense.

The adversary's objective can be expressed as an optimization problem. For example, the following
constrained maximization of the loss function is one way of formulating how an adversary could
generate an untargeted adversarial input $x^{adv}$.\nopagebreak

\begin{alignat*}{4}
  \Delta x = &\argmax_{\delta}  && J(x + \delta, y) \\
  &\text{subject to} && \  \|\delta\|_p \leq \epsilon \\
  &                  && x + \delta \in \mathcal{X}
\end{alignat*}

There are various ways to generate attacks. Under many formulations it's challenging to devise an
exact computation of $\Delta x$ that optimizes the objective function. An approximation is often
employed.

\textbf{Fast Gradient Sign Method~(FGSM)}~\cite{goodfellow_explaining_2015} generates an adversarial
perturbation $\Delta x$ = $\epsilon \cdot \sign(\nabla_x J(x, y))$, which is the approximate
direction of the loss function gradient. The $\sign$ function bounds its input to an $L_\infty$ norm
of 1, which is scaled \mbox{by $\epsilon$}.

\textbf{Basic Iterative Method~(BIM)}~\cite{kurakin_adversarial_2017} iteratively applies FGSM,
whereby $x^{adv}_{t} = x^{adv}_{t-1} + \alpha \cdot \sign(\nabla_x J(x^{adv}_{t-1}, y))$ for each
step, starting with $x^{adv}_0 = x$. The $L_\infty$ norm is bounded by $\alpha$ on each iteration
and by $t\cdot\alpha$ after $t$ iterations. $x^{adv}_t$ can be clipped after each iteration in a way
that constrains the final $x^{adv}$ to an $\epsilon$-ball of $x$.

\textbf{Carlini \& Wagner (CW)}~\cite{carlini_towards_2017} generates an adversarial perturbation
via gradient descent to solve $\Delta x = \argmin_{\delta} (\|\delta\|_p + c \cdot f(x + \delta))$
subject to a box constraint on $x + \delta$. $f$ is a function for which $f(x + \delta) \leq 0$ if
and only if the target classifier is successfully attacked. Experimentation yielded the most
effective $f$---for targeted attacks---of those considered. $c$ is a positive constant that can be
found with binary search, a strategy that worked well empirically. Clipping or a change of variables
can be used to accommodate the box constraint.

\subsection{Ensembling}

Our research draws inspiration from ensembling, the combination of multiple models to improve
performance relative to the component models themselves. There are various ways of combining models.
An approach that is widely used in deep learning averages outputs from an assortment of neural
networks; each network having the same architecture, trained from a differing set of randomly
initialized weights.

\section{Method}
\label{sec:method}

To detect adversarial instances, we use hidden layer representations---from \emph{representation
  models}---as inputs to adversarial \emph{detection models}. For our experiments in
Section~\ref{sec:experiments}, the representation models are convolutional neural networks that are
independently trained for the same classification task, initialized with different weights.
Representations are extracted from the penultimate layers of the trained networks. The method we
describe in this section is more general, as various approaches could be used for preparing
representation models. For example, each representation model could be an independently trained
autoencoder---as opposed to a classifier---with representations for each model extracted from
arbitrary hidden layers. Additionally, it's not necessary that each of the models---used for
extracting representations---has the same architecture.

We devise two broad techniques---\emph{model-wise} and \emph{unit-wise}---for extracting
representations and detecting adversarial instances. These approaches each have two formulations, a
\emph{treatment} that incorporates multiple representation models and a \emph{control} that uses a
single representation model. For each technique, the functional form of the detection step is the
same across treatment and control. This serves our objective of measuring the contribution of
incrementally incorporating multiple representation models, as the control makes it possible to
check whether gains are coming from some aspect other than the incorporation of multiple
representation models.

The illustrations in this section are best viewed in color.

\subsection{Model-Wise Detection}

With $N$ representation models, model-wise detection uses a set of representations from each
underlying model as separate input to $N$ corresponding detection models that each outputs an
adversarial score. These scores, which we interpret as estimated probabilities, are then averaged to
give an ensemble adversarial probability estimate. A baseline---holding fixed the number of
detectors---uses a single representation model as a repeated input to multiple detection models. The
steps of both approaches are outlined below.

\subsubsection{Model-Wise Treatment}

\paragraph{Step 1} Extract representations for input $x$ from $N$ representation models.
\begin{center}
  \begin{tabular}{cccc}
    \includegraphics[height=1.2cm]{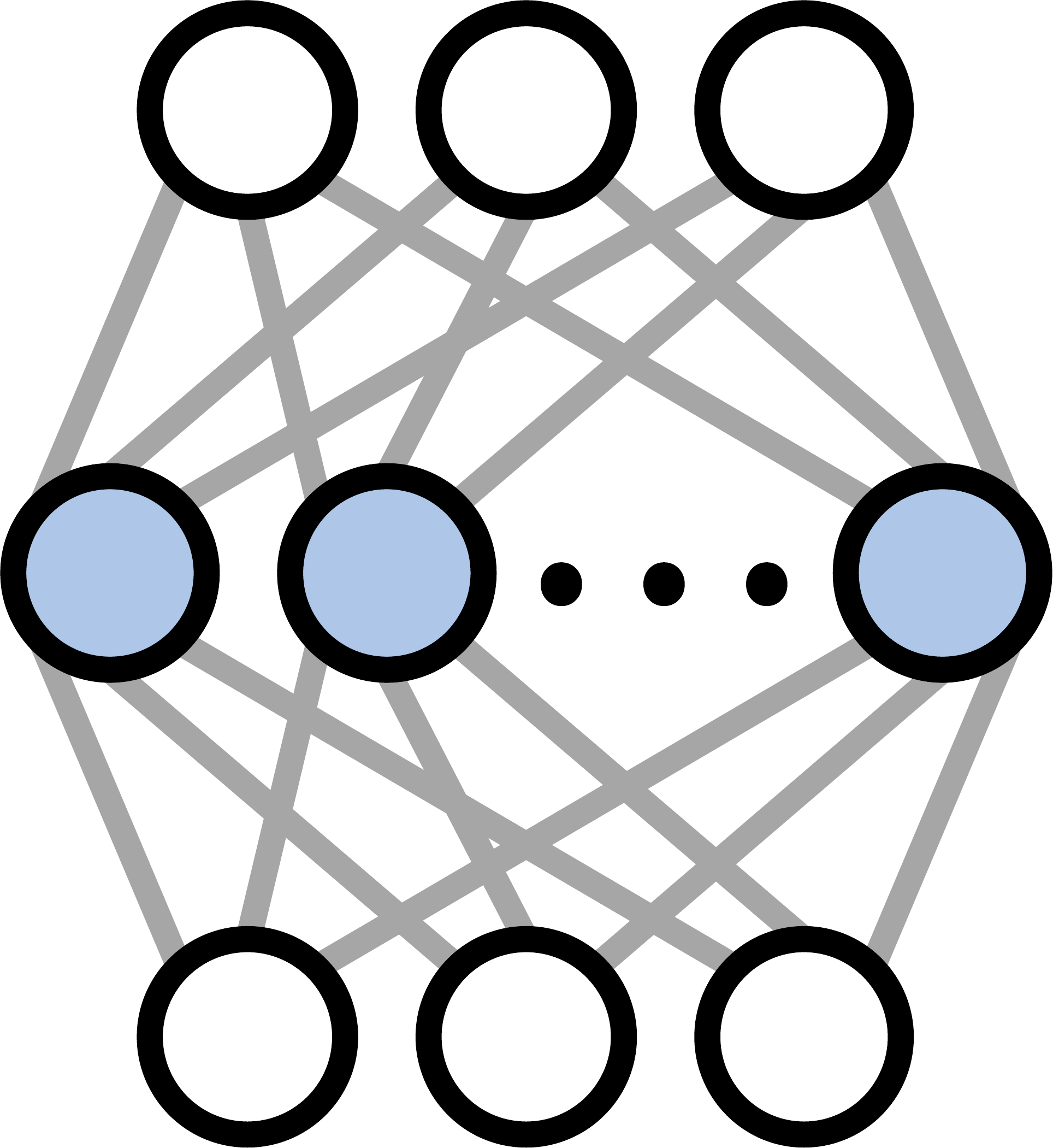} &
    \includegraphics[height=1.2cm]{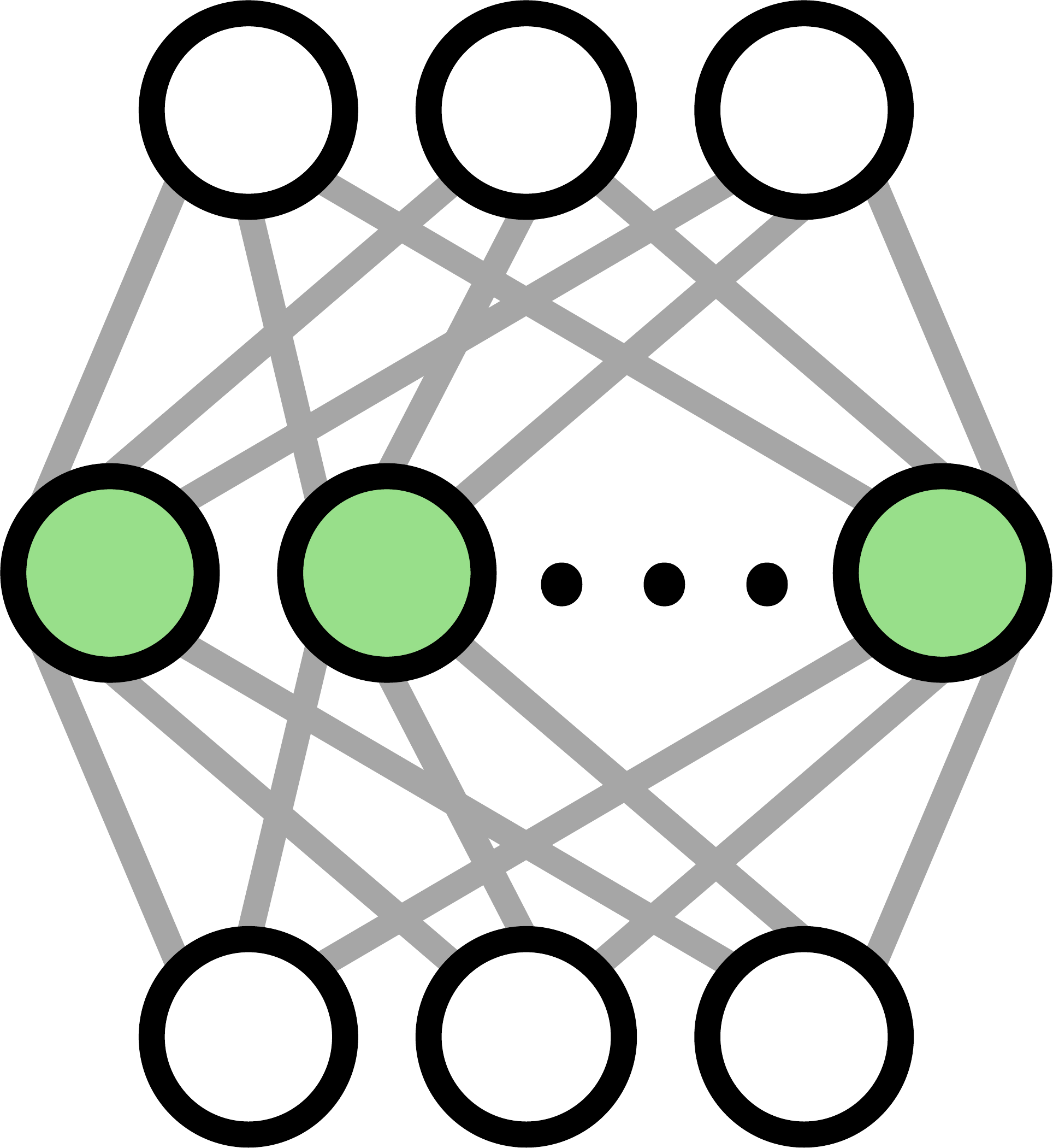} &
    \multirow[b]{1}{*}[15pt]{\begin{tabular}{@{}c@{}}\huge...\end{tabular}} &
    \includegraphics[height=1.2cm]{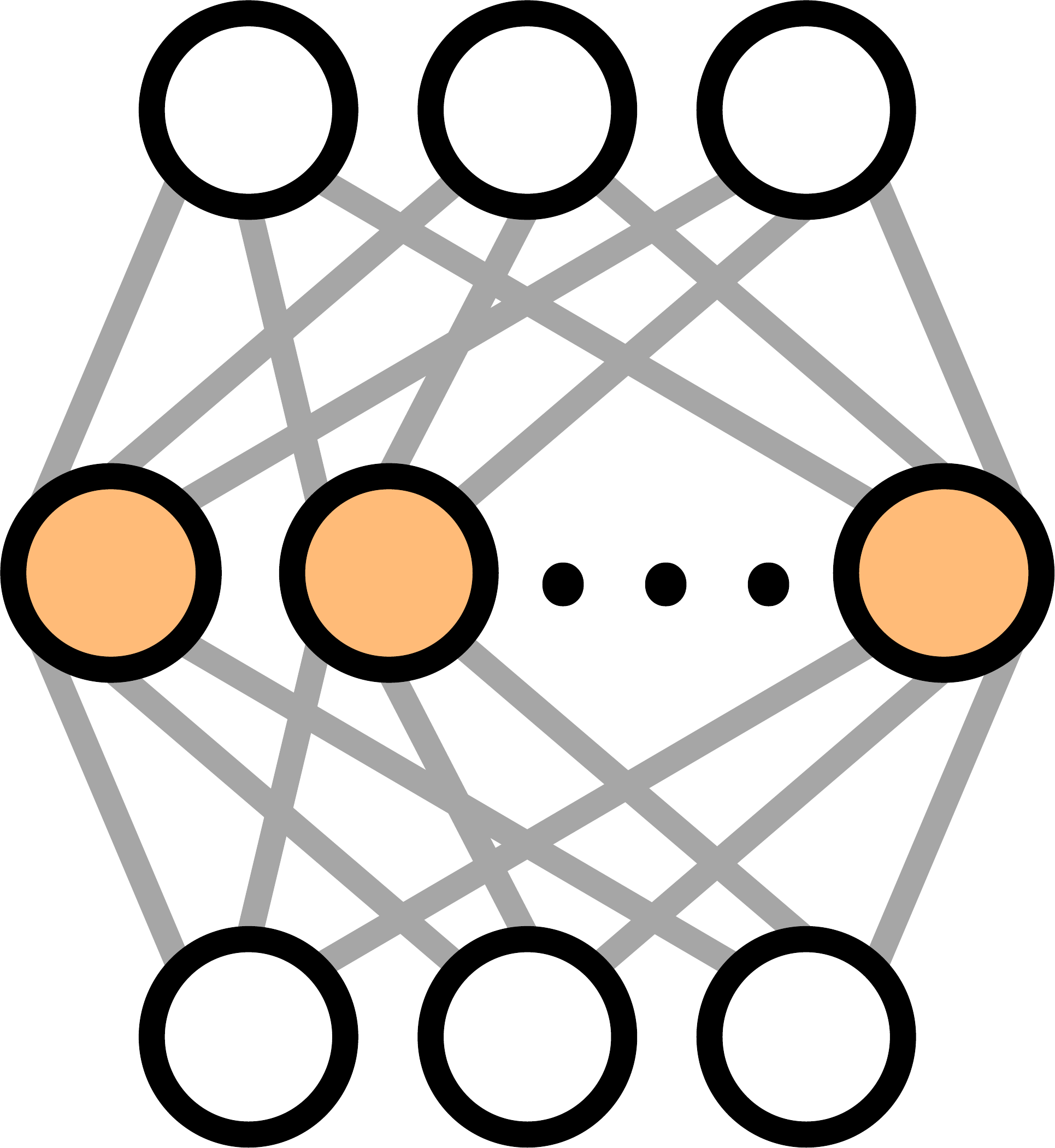} \\
    $x$ & $x$ & & $x$
  \end{tabular}
\end{center}

\paragraph{Step 2} Pass the \emph{Step 1} representations through $N$ corresponding detection models
that each output adversarial probability (denoted $P_i$ for model~$i$).
\begin{center}
  \begin{tabular}{cccc}
    $P_1$ & $P_2$ & & $P_N$ \\
    \includegraphics[height=1.2cm]{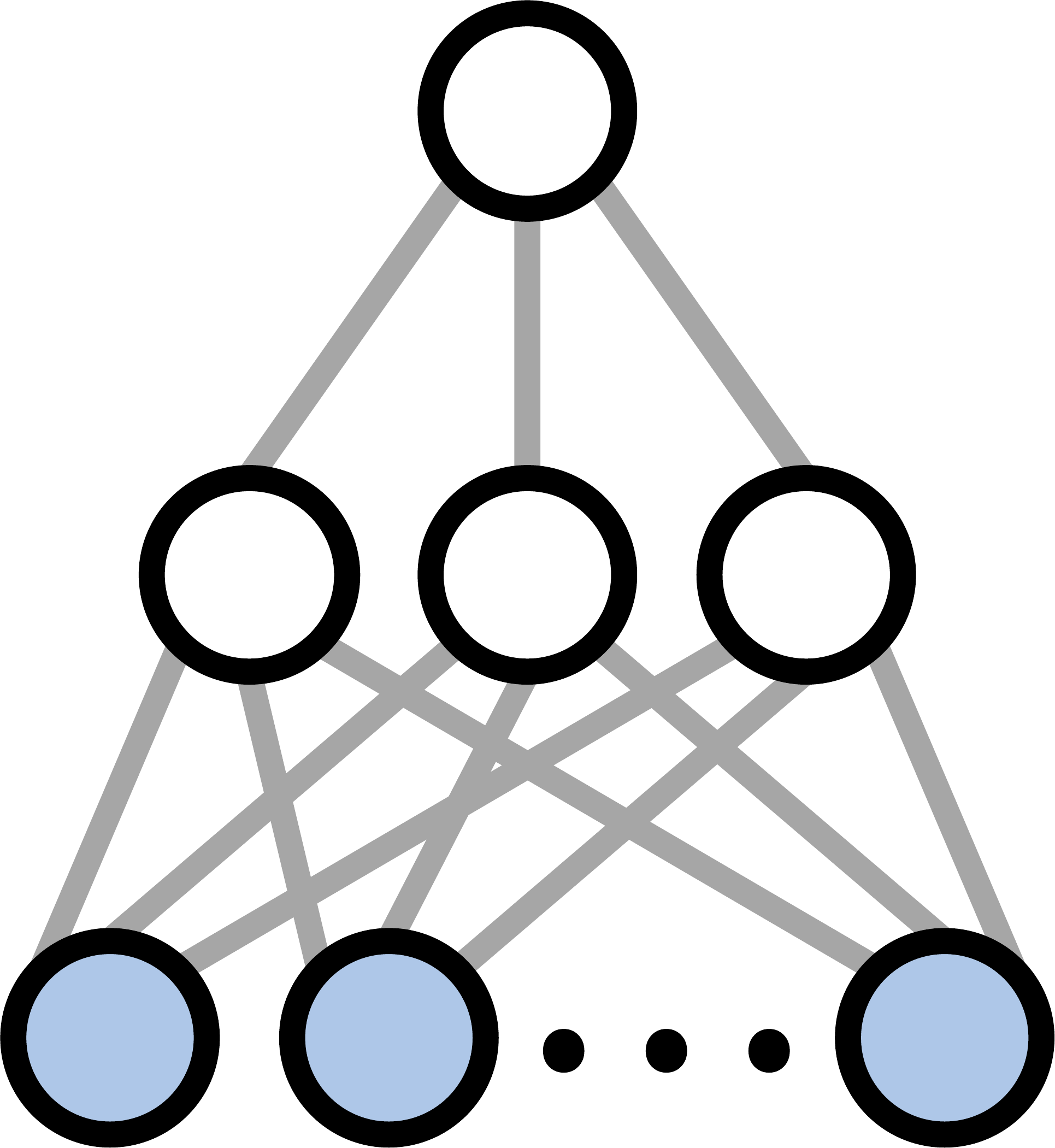} &
    \includegraphics[height=1.2cm]{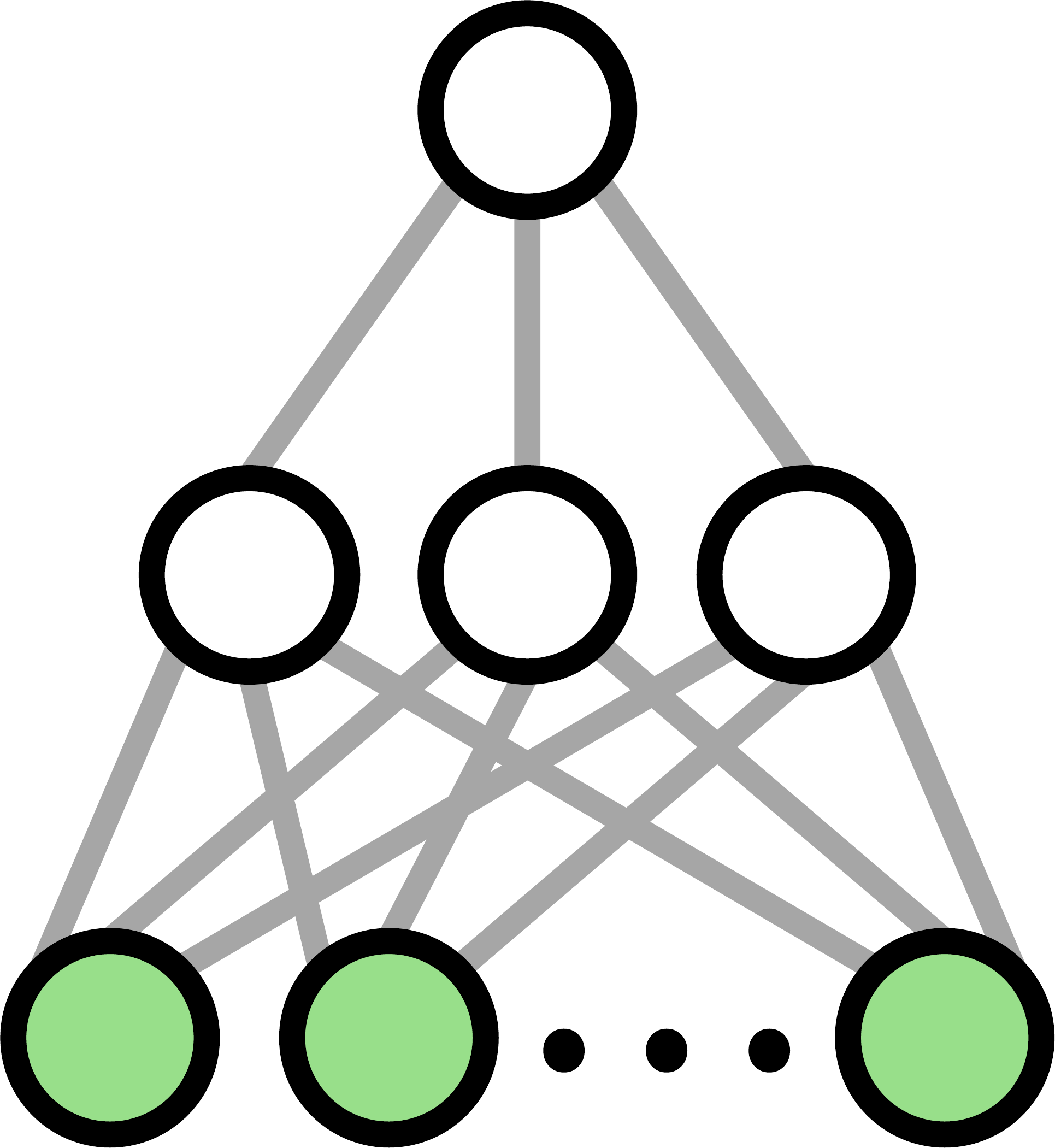} &
    \multirow[b]{1}{*}[15pt]{\begin{tabular}{@{}c@{}}\huge...\end{tabular}} &
    \includegraphics[height=1.2cm]{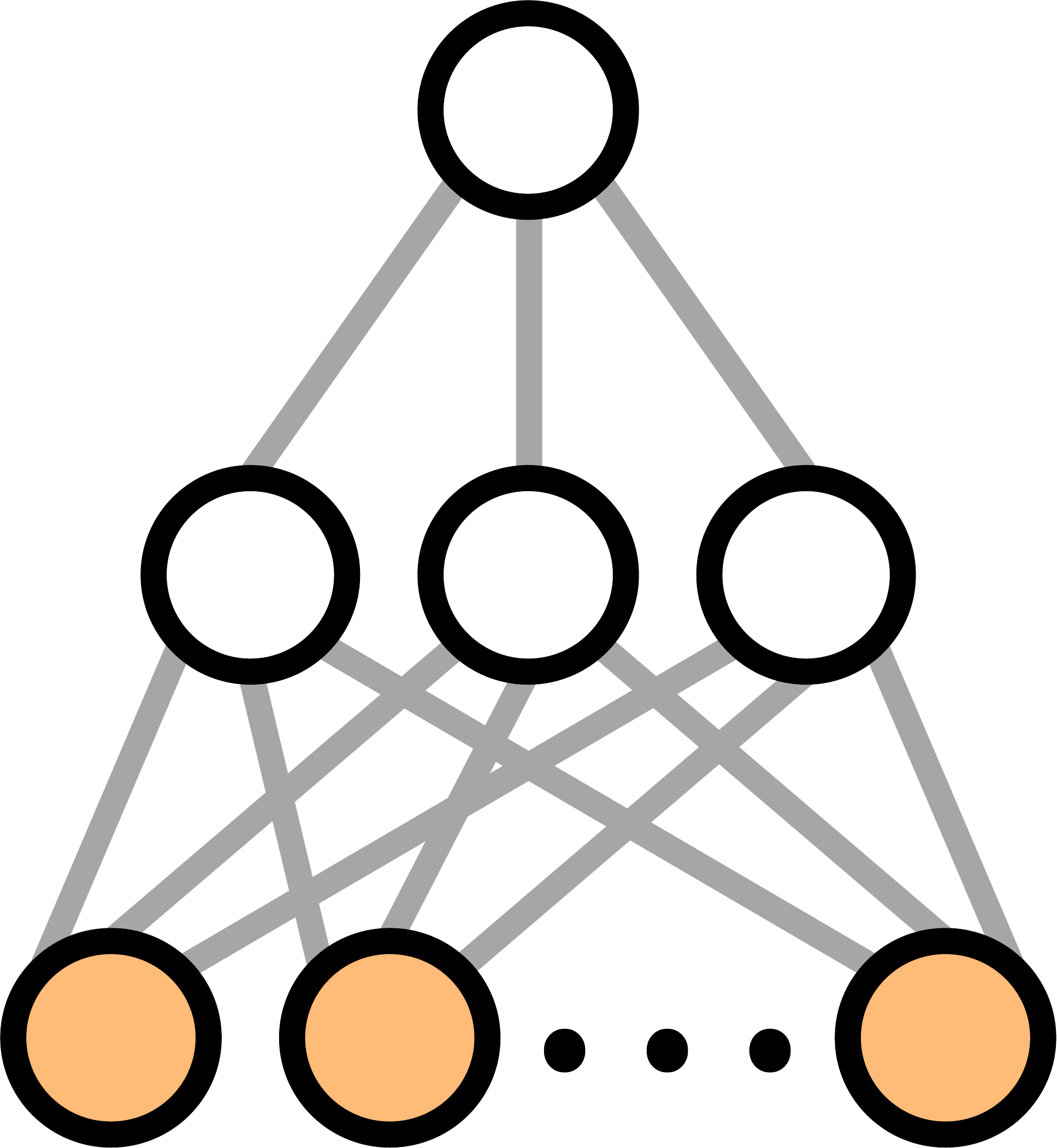}
  \end{tabular}
\end{center}

\paragraph{Step 3} Calculate adversarial probability $P$ as the average of \emph{Step 2} adversarial
probabilities.
\begin{equation*}
  P = \frac{1}{N}\sum_{i=1}^{N}{P_i}
\end{equation*}

\subsubsection{Model-Wise Control}

\paragraph{Step 1} Extract representations for input $x$ from a single representation model.
\begin{center}
  \begin{tabular}{c}
    \includegraphics[height=1.2cm]{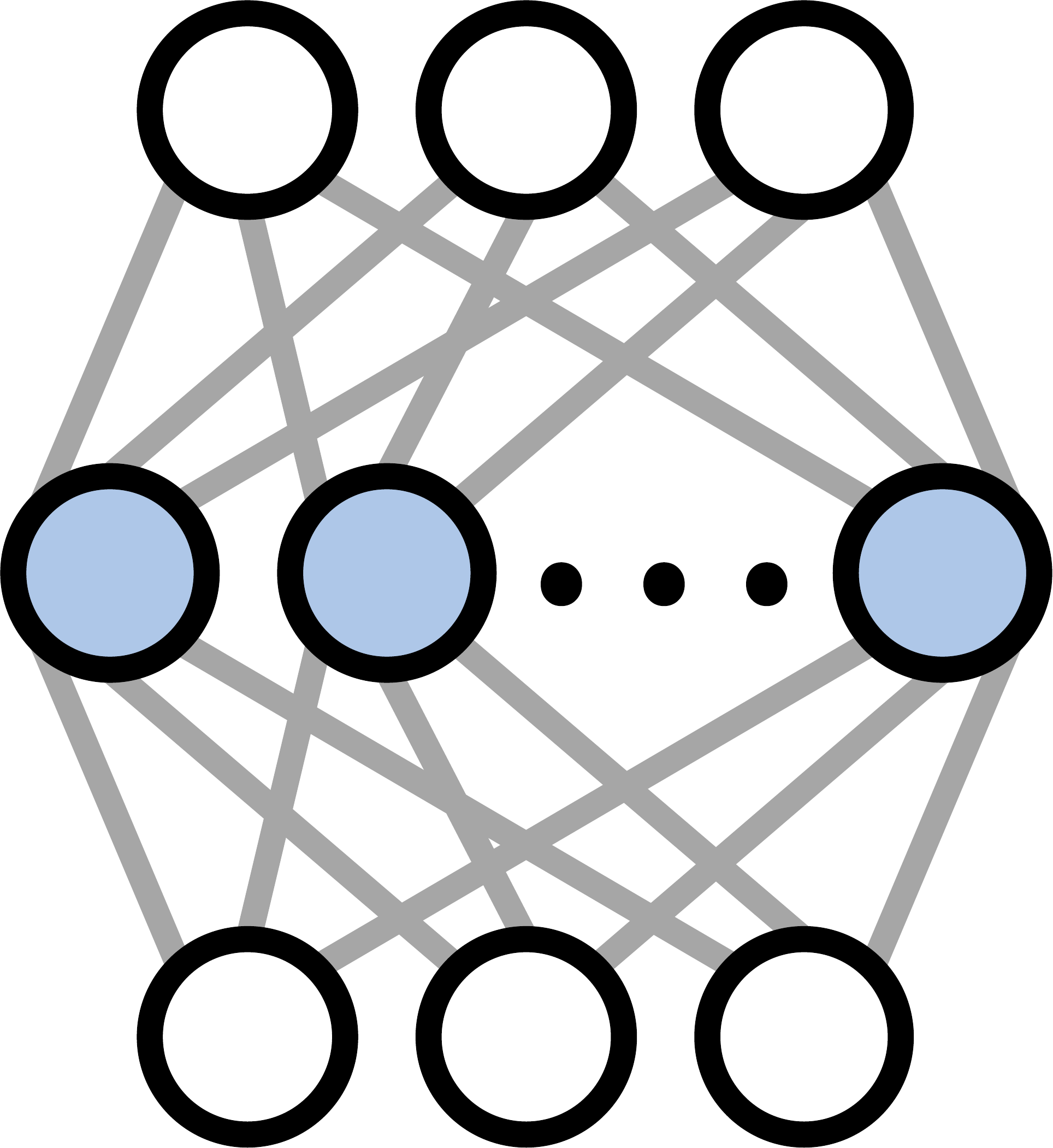} \\
    $x$
  \end{tabular}
\end{center}

\paragraph{Step 2} Pass the \emph{Step 1} representations through $N$ detection models that each
  outputs adversarial probability (denoted $P_i$ for model~$i$).
\begin{center}
  \begin{tabular}{cccc}
    $P_1$ & $P_2$ & & $P_N$ \\
    \includegraphics[height=1.2cm]{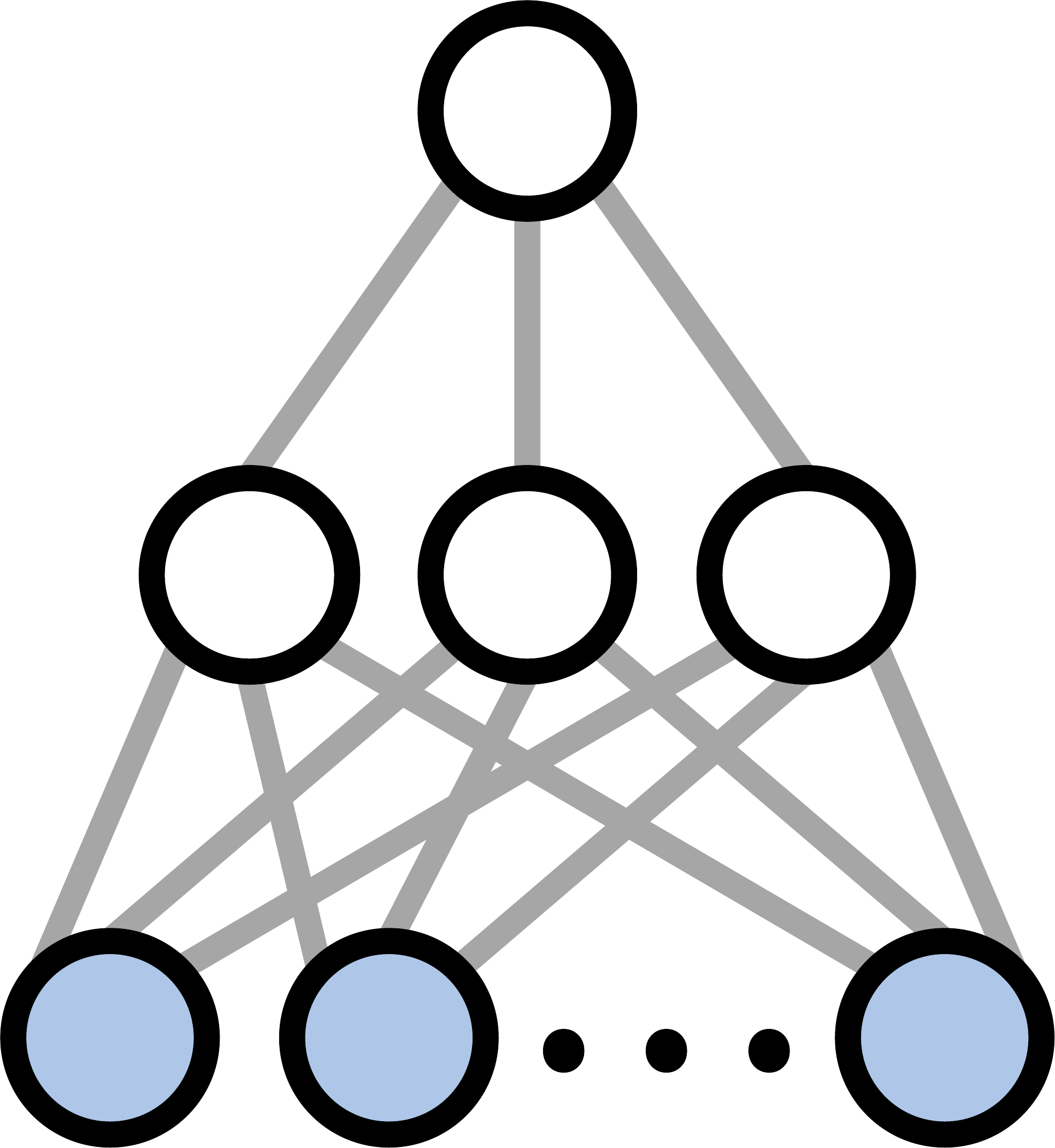} &
    \includegraphics[height=1.2cm]{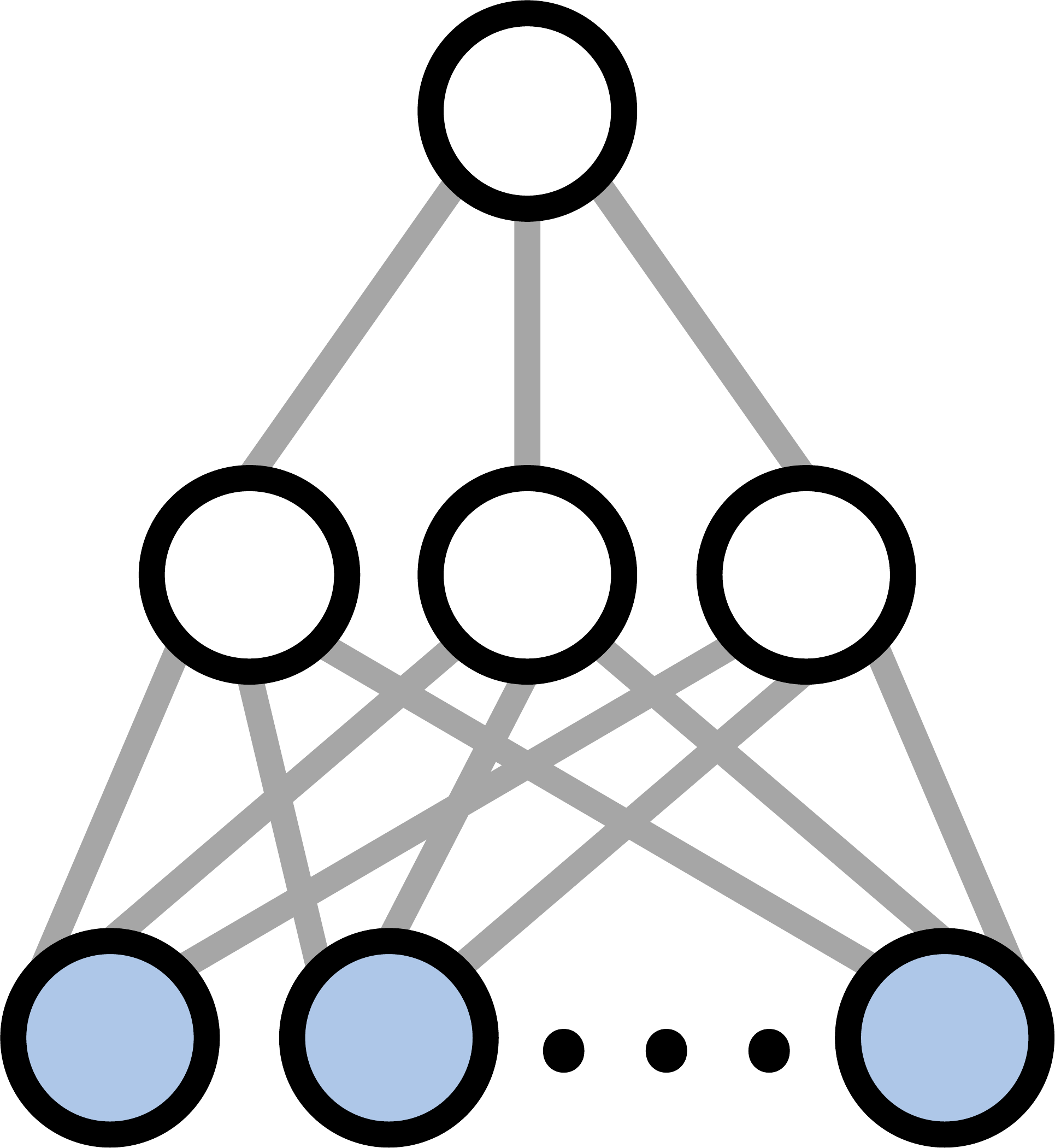} &
    \multirow[b]{1}{*}[15pt]{\begin{tabular}{@{}c@{}}\huge...\end{tabular}} &
    \includegraphics[height=1.2cm]{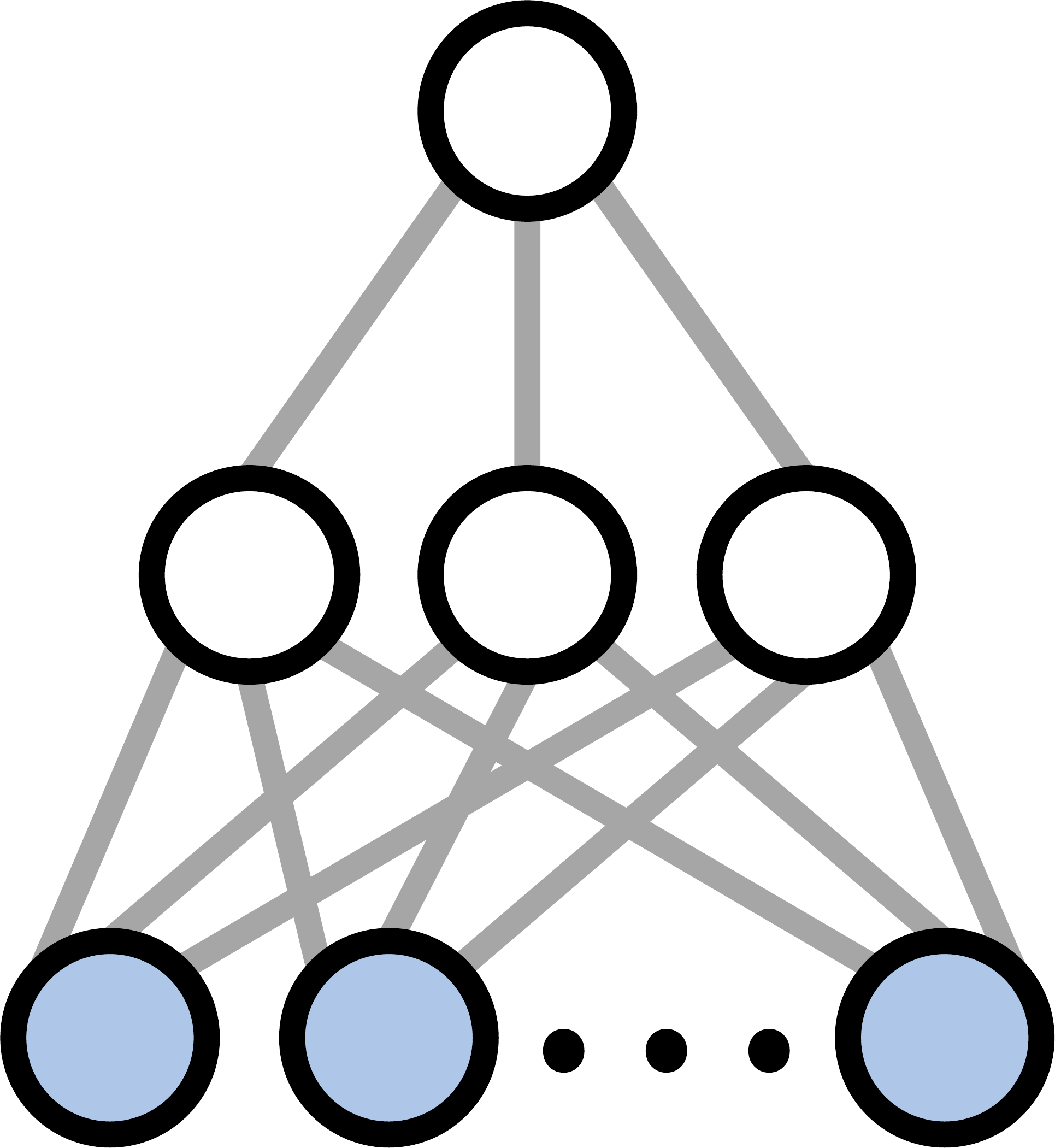}
  \end{tabular}
\end{center}

\paragraph{Step 3} Calculate adversarial probability $P$ as the average of \emph{Step 2}
adversarial probabilities.
\begin{equation*}
  P = \frac{1}{N}\sum_{i=1}^{N}{P_i}
\end{equation*}

\subsection{Unit-Wise Detection}

With $N$ representation models, unit-wise detection incorporates a single representation from each
underlying model to form an $N$-dimensional array of features as input to a single detection model.
A baseline---holding fixed the number of features for the detector---uses a set of units from a
single representation model to form an input array for a detection model. The steps of both
approaches are outlined below.

\subsubsection{Unit-Wise Treatment}

\begin{samepage}
\paragraph{Step 1} Extract a single representation for input $x$ from $N$ representation models.
There is no requirement on which unit is selected nor whether there is any correspondence between
which unit is selected from each model.

\begin{center}
  \begin{tabular}{cccc}
    \includegraphics[height=1.2cm]{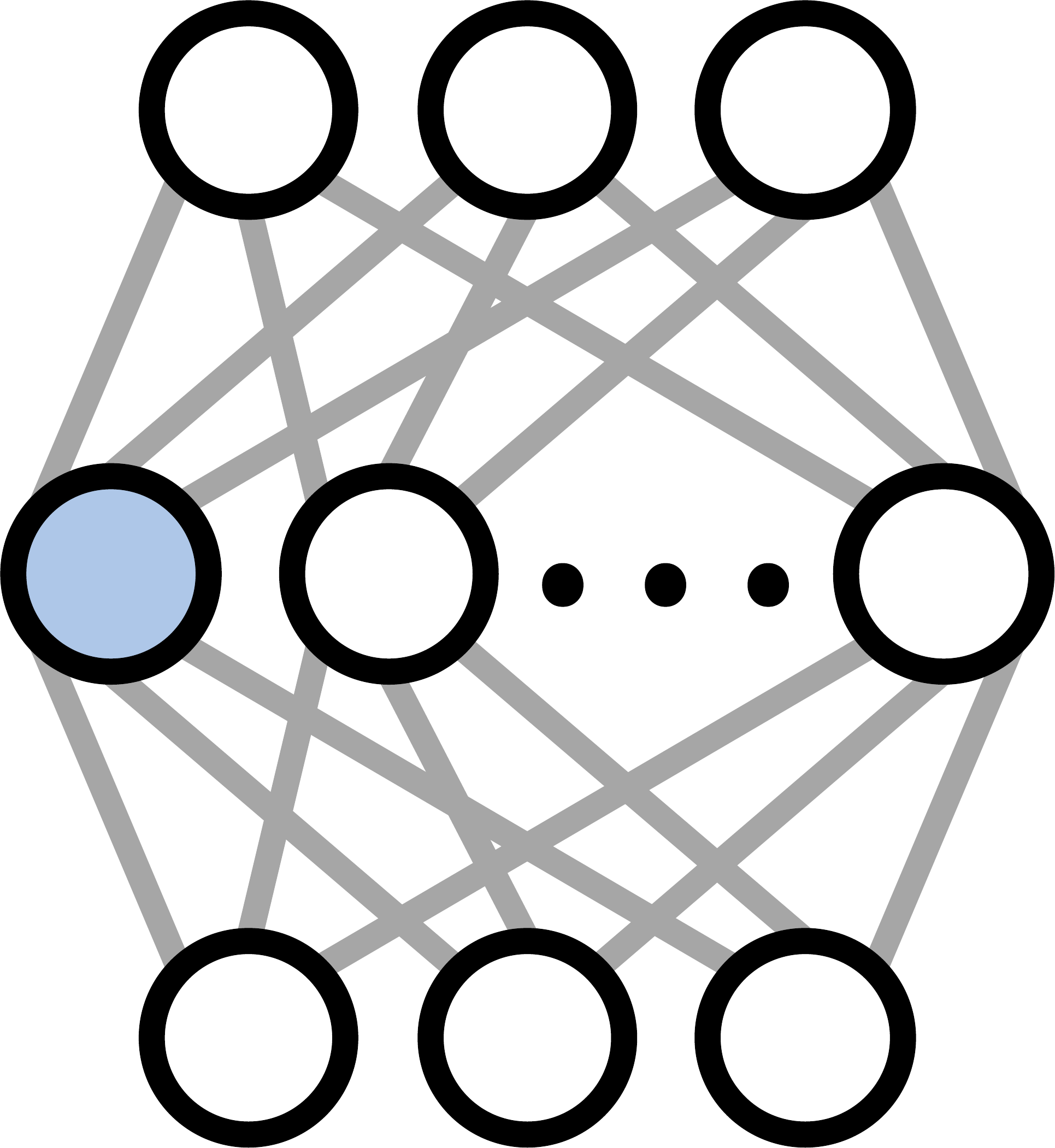} &
    \includegraphics[height=1.2cm]{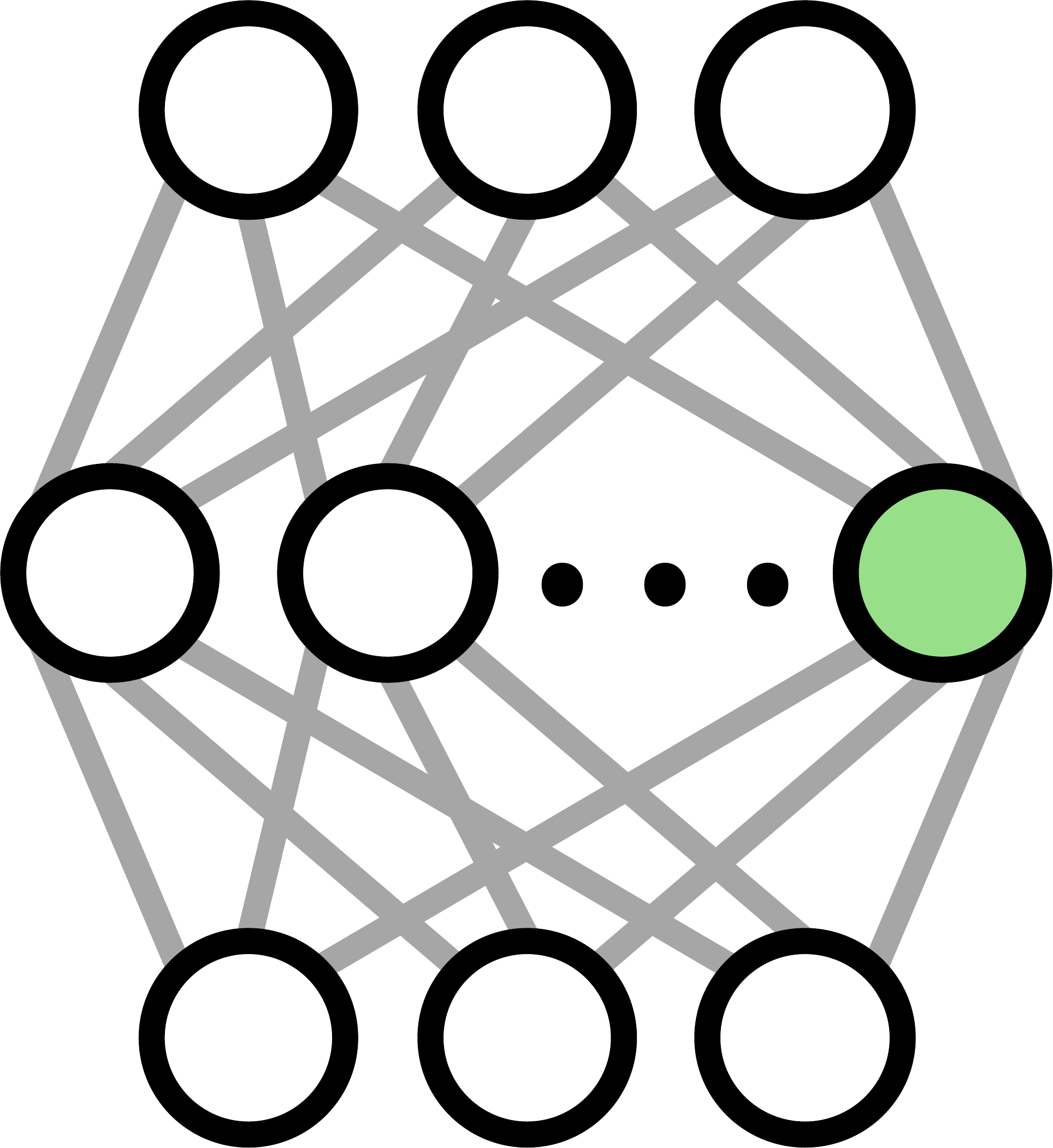} &
    \multirow[b]{1}{*}[15pt]{\begin{tabular}{@{}c@{}}\huge...\end{tabular}} &
    \includegraphics[height=1.2cm]{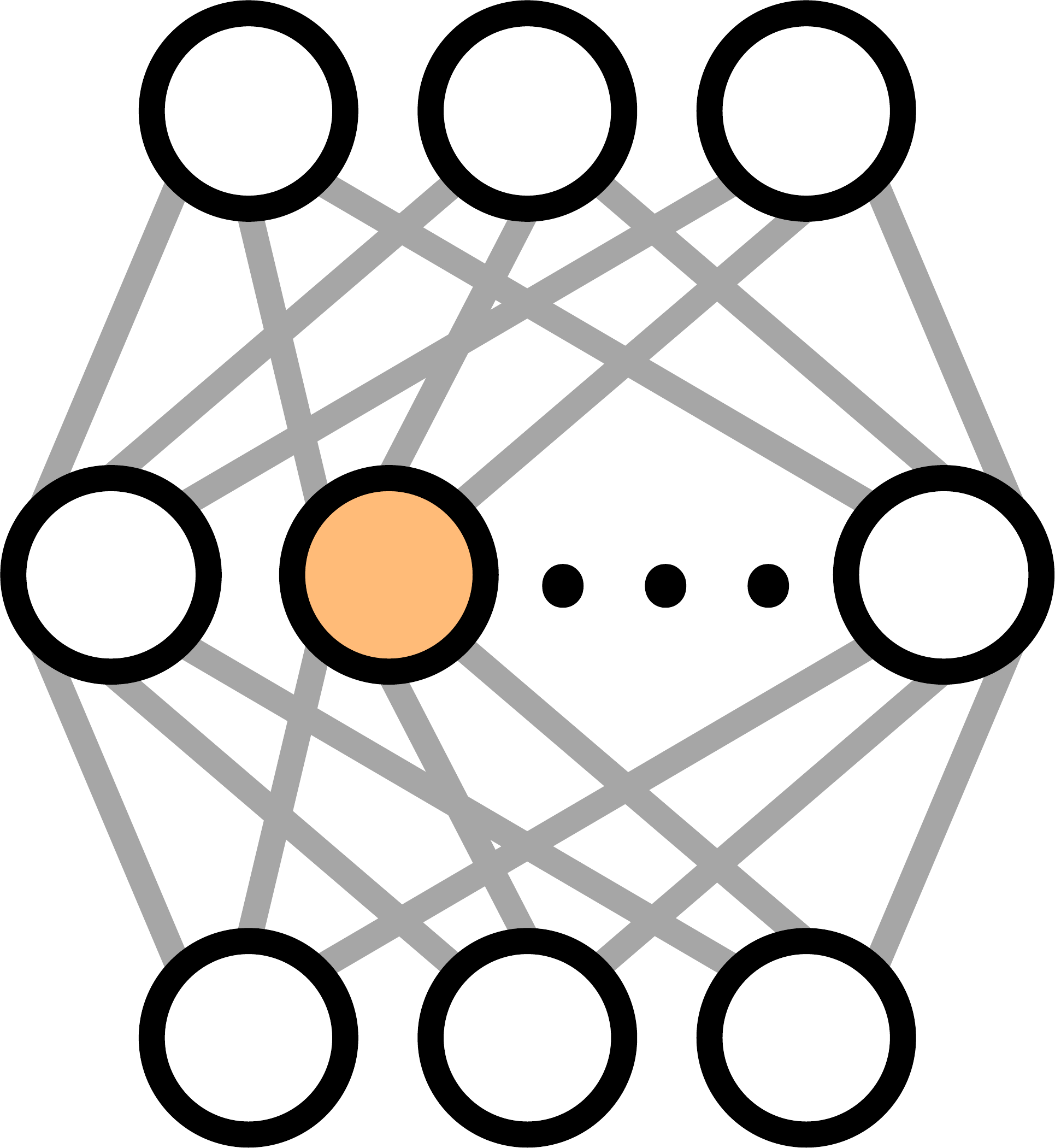} \\
    $x$ & $x$ & & $x$
  \end{tabular}
\end{center}
\end{samepage}

\begin{samepage}
\paragraph{Step 2} Pass the $N$-dimensional array of \emph{Step 1} representations through an
adversarial detection model that outputs adversarial probability $P$.
\begin{center}
  \begin{tabular}{c}
    $P$ \\
    \includegraphics[height=1.2cm]{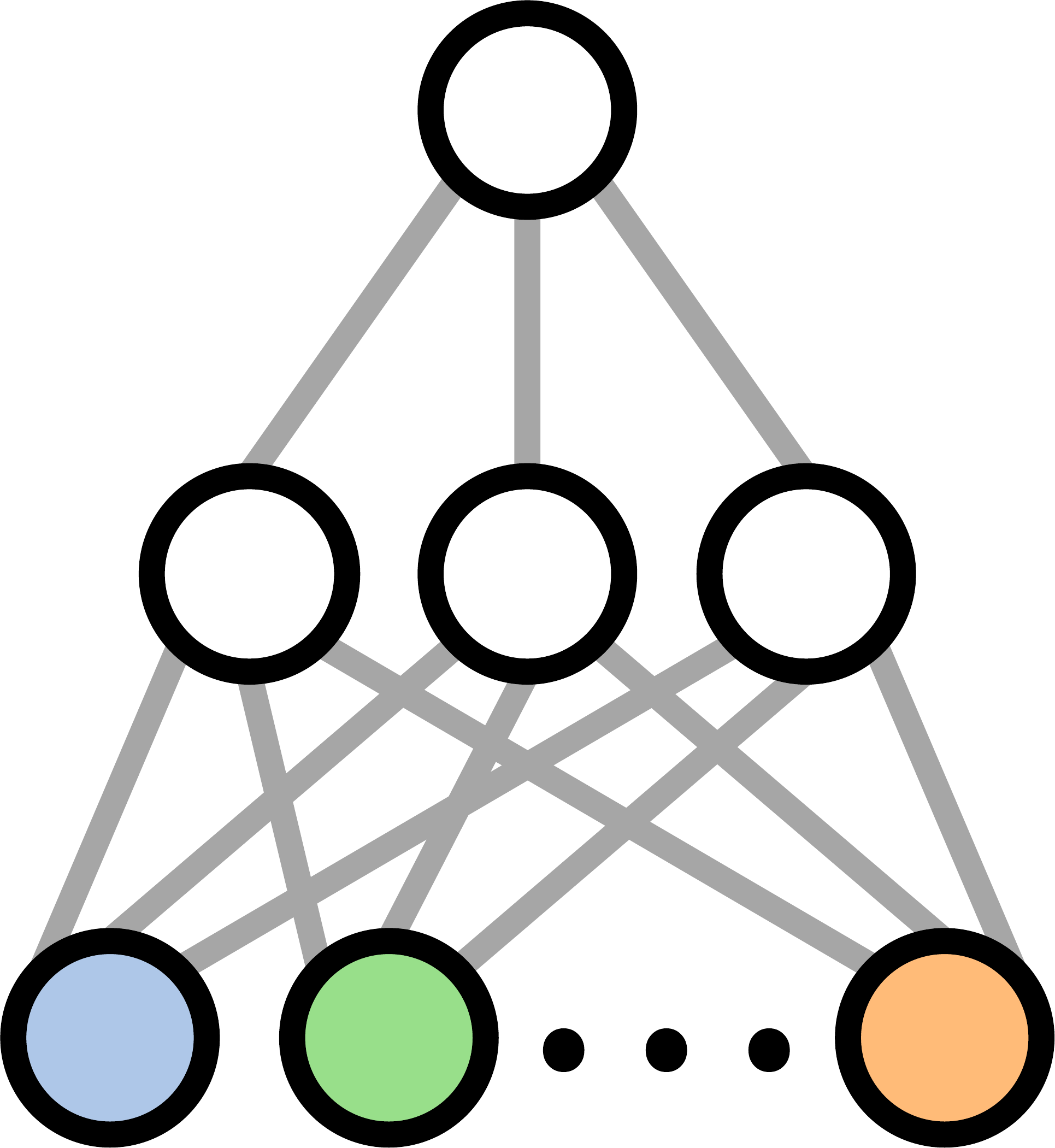}
  \end{tabular}
\end{center}
\end{samepage}

\subsubsection{Unit-Wise Control}

\begin{samepage}
\paragraph{Step 1} Extract $N$ units from the representations for input $x$ from a single
representation model. In the illustration that follows, the count of extracted representation units,
$N$, matches the total number of units available. It's also possible for $N$ to be smaller than the
quantity available.
\begin{center}
  \begin{tabular}{c}
    \includegraphics[height=1.2cm]{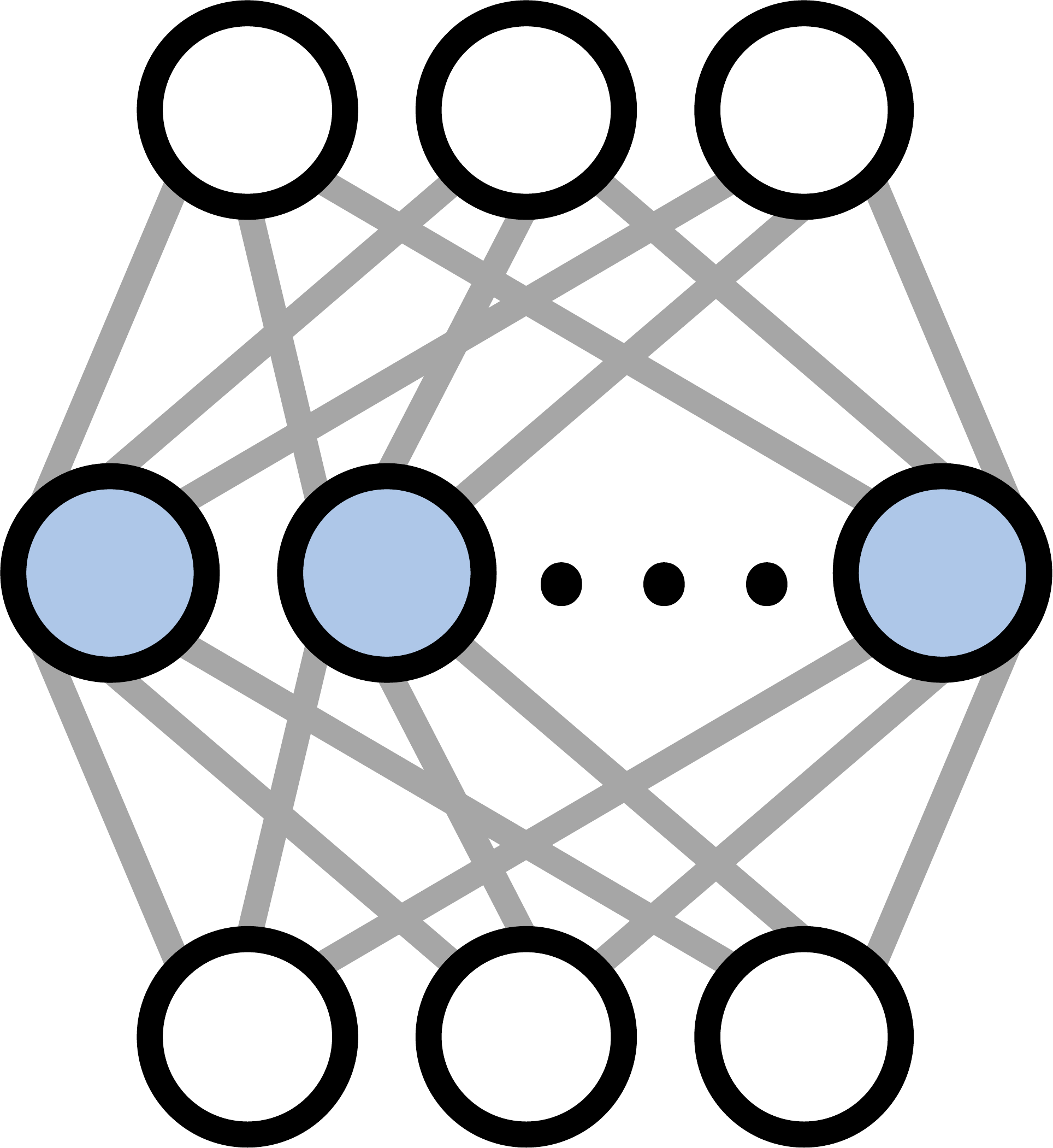} \\
    $x$ \\
  \end{tabular}
\end{center}
\end{samepage}

\begin{samepage}
\paragraph{Step 2} Pass \emph{Step 1} representations through an adversarial detection model that
  outputs adversarial probability $P$.
\begin{center}
  \begin{tabular}{c}
    $P$ \\
    \includegraphics[height=1.2cm]{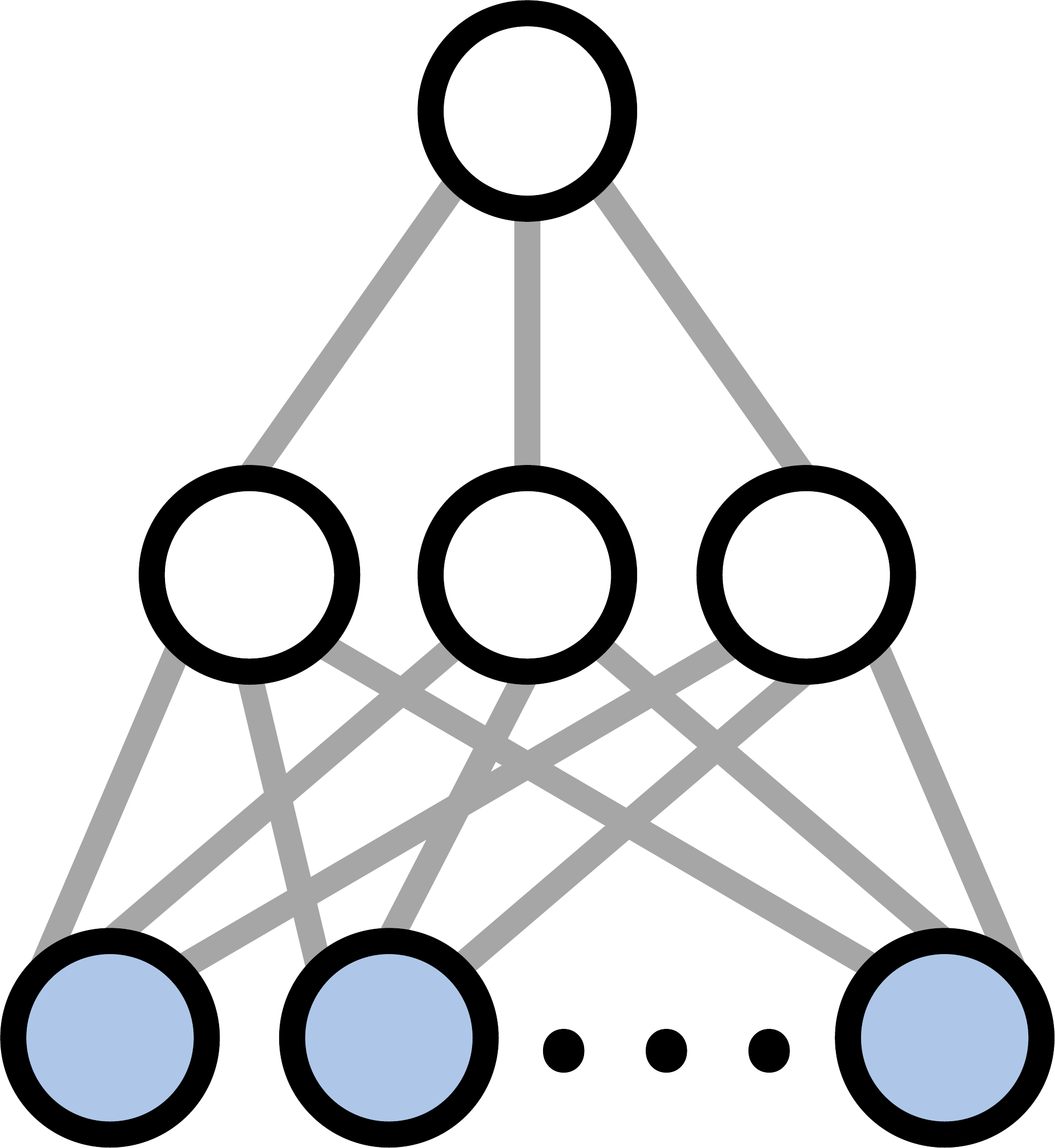}
  \end{tabular}
\end{center}
\end{samepage}

\subsection{Measuring the Contribution from Multiple Models}

We are interested in measuring the contribution of multiple models for detecting adversarial
instances. For both the model-wise and unit-wise detection techniques, the contribution of multiple
models can be evaluated by inspecting the change in treatment performance when incrementing the
number of representation models, $N$. The changes should be considered relative to the control
performance, to check whether any differences are coming from some aspect other than the
incorporation of multiple representation models.

\section{Experiments}
\label{sec:experiments}

\subsection{Experimental Settings}

We conducted experiments using the CIFAR-10 dataset~\cite{krizhevsky_learning_2009}, which is
comprised of 60,000 $32{\times}32$ RGB images across 10 classes. The dataset, as received, was
already split into 50,000 training images and 10,000 test images. We trained one neural network
classifier that served as the target for generating adversarial attacks. We trained 1,024 additional
neural network classifiers to be used as representation models---with representations extracted from
the 512-dimensional penultimate layer of each network. A different randomization seed was used for
initializing the weights of the 1,025 networks. Each network had the same---18-layer,
11,173,962-parameter---ResNet-inspired architecture, with filter counts and depth matching
the~\citeauthor{kuangliu_kuangliupytorch-cifar_2021} ResNet-18 architecture.\footnote{This differs
from the ResNet-20 architecture used for CIFAR-10 in the original ResNet paper~\cite{he_deep_2016}.}
Pixel values of input images were scaled by $1/255$ to be between 0 and 1. The networks were trained
for 100 epochs using an Adam optimizer \cite{kingma_adam:_2014}, with random horizontal flipping and
random crop sampling on images padded with 4 pixels per edge. The model for attack generation had
91.95\% accuracy on the test dataset. The average test accuracy across the 1,024 additional networks
was 92.22\% with sample standard deviation of 0.34\%.

\subsubsection{Adversarial Attacks}

Untargeted adversarial perturbations were generated for the 9,195 images that were originally
correctly classified by the attacked model. Attacks were conducted with FGSM, BIM, and CW, all using
the \texttt{cleverhans} library~\cite{papernot2018cleverhans}. After each attack, we clipped the
perturbed images between 0 and 1 and quantized the pixel intensities to 256 discrete values. This
way the perturbed instances could be represented in 24-bit RGB space.

For FGSM, we set $\epsilon = 3 / 255$ for a maximum perturbation of 3 intensity values (out of 255)
for each pixel on the unnormalized data. Model accuracy on the attacked model---for the 9,195
perturbed images---was 21.13\% (i.e., an attack success rate of 78.87\%). Average accuracy on the
1,024 representation models was 61.69\% (i.e., an attack transfer success rate of 38.31\%) with
sample standard deviation of 1.31\%.

For BIM, we used 10 iterations with $\alpha = 1 / 255$ and maximum perturbation magnitude clipped to
$\epsilon = 3 / 255$. This results in a maximum perturbation of 1 unnormalized intensity value per
pixel on each step, with maximum perturbation after all steps clipped to 3. Accuracy after attack
was 0.61\% for the attacked model. Average accuracy on the 1,024 representation models was 41.09\%
with sample standard deviation of 2.64\%.

For CW, we used an $L_2$ norm distance metric along with most default parameters---a learning rate
of 0.005, 5 binary search steps, and 1,000 maximum iterations. We raised the confidence
parameter\footnote{Our description of CW in Section~\ref{sec:preliminaries} does not discuss the
  $\kappa$ confidence parameter. See the CW paper~\cite{carlini_towards_2017} for details.} to 100
from its default of 0, which increases attack transferability. This makes our experiments more
closely align with black-box and grey-box attack scenarios, where transferability would be an
objective of an adversary. Accuracy after attack was 0.07\% for the attacked model. Average accuracy
on the 1,024 representation models was 5.86\% with sample standard deviation of 1.72\%.

\begin{figure}[tb]
  \begin{center}
    {
      \renewcommand{\arraystretch}{2.2}
      \newcommand\imgwidth{0.095\columnwidth}
      \newcommand\colwidth{1.15cm}
      \begin{tabular}{
          r>{\centering\arraybackslash}p{\colwidth}
          >{\centering\arraybackslash}p{\colwidth}
          >{\centering\arraybackslash}p{\colwidth}
          >{\centering\arraybackslash}p{\colwidth}}
        & Original & FGSM & BIM & CW \\
        \addlinespace[-1ex]  % Reduce space between header and first row
        airplane &
        \includegraphics[align=c,width=\imgwidth]{%
          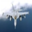} &
        \includegraphics[align=c,width=\imgwidth]{%
          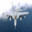} &
        \includegraphics[align=c,width=\imgwidth]{%
          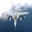} &
        \includegraphics[align=c,width=\imgwidth]{%
          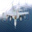} \\
        automobile &
        \includegraphics[align=c,width=\imgwidth]{%
          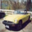} &
        \includegraphics[align=c,width=\imgwidth]{%
          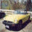} &
        \includegraphics[align=c,width=\imgwidth]{%
          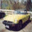} &
        \includegraphics[align=c,width=\imgwidth]{%
          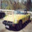} \\
        bird &
        \includegraphics[align=c,width=\imgwidth]{%
          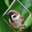} &
        \includegraphics[align=c,width=\imgwidth]{%
          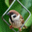} &
        \includegraphics[align=c,width=\imgwidth]{%
          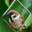} &
        \includegraphics[align=c,width=\imgwidth]{%
          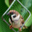} \\
        cat &
        \includegraphics[align=c,width=\imgwidth]{%
          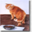} &
        \includegraphics[align=c,width=\imgwidth]{%
          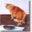} &
        \includegraphics[align=c,width=\imgwidth]{%
          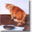} &
        \includegraphics[align=c,width=\imgwidth]{%
          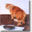} \\
        deer &
        \includegraphics[align=c,width=\imgwidth]{%
          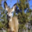} &
        \includegraphics[align=c,width=\imgwidth]{%
          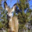} &
        \includegraphics[align=c,width=\imgwidth]{%
          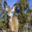} &
        \includegraphics[align=c,width=\imgwidth]{%
          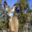} \\
        dog &
        \includegraphics[align=c,width=\imgwidth]{%
          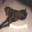} &
        \includegraphics[align=c,width=\imgwidth]{%
          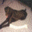} &
        \includegraphics[align=c,width=\imgwidth]{%
          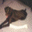} &
        \includegraphics[align=c,width=\imgwidth]{%
          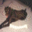} \\
        frog &
        \includegraphics[align=c,width=\imgwidth]{%
          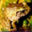} &
        \includegraphics[align=c,width=\imgwidth]{%
          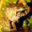} &
        \includegraphics[align=c,width=\imgwidth]{%
          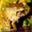} &
        \includegraphics[align=c,width=\imgwidth]{%
          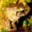} \\
        horse &
        \includegraphics[align=c,width=\imgwidth]{%
          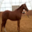} &
        \includegraphics[align=c,width=\imgwidth]{%
          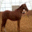} &
        \includegraphics[align=c,width=\imgwidth]{%
          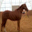} &
        \includegraphics[align=c,width=\imgwidth]{%
          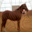} \\
        ship &
        \includegraphics[align=c,width=\imgwidth]{%
          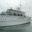} &
        \includegraphics[align=c,width=\imgwidth]{%
          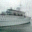} &
        \includegraphics[align=c,width=\imgwidth]{%
          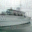} &
        \includegraphics[align=c,width=\imgwidth]{%
          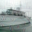} \\
        truck &
        \includegraphics[align=c,width=\imgwidth]{%
          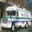} &
        \includegraphics[align=c,width=\imgwidth]{%
          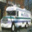} &
        \includegraphics[align=c,width=\imgwidth]{%
          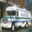} &
        \includegraphics[align=c,width=\imgwidth]{%
          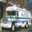}
      \end{tabular}
    }
  \end{center}
  \caption{
    Example CIFAR-10 images after adversarial perturbation. The original image---in the leftmost
    column---is followed by three columns corresponding to FGSM, BIM, and CW attacks, respectively.
    Images were chosen randomly from the set of test images that were correctly classified without
    perturbation---the population of images for which attacks were generated.
  }
  \label{fig:attacked_images}
\end{figure}

Figure~\ref{fig:attacked_images} shows examples of images that were perturbed for our experiments.
These were chosen randomly from the 9,195 correctly classified test images---the population of
images for which attacks were generated.

\subsubsection{Adversarial Detectors}

% This figure was moved here for better layout. See <original>fig:model_wise for the original
% location in the LaTeX markup.
\begin{figure*}[t]
  \centering
  \includegraphics[width=\linewidth]{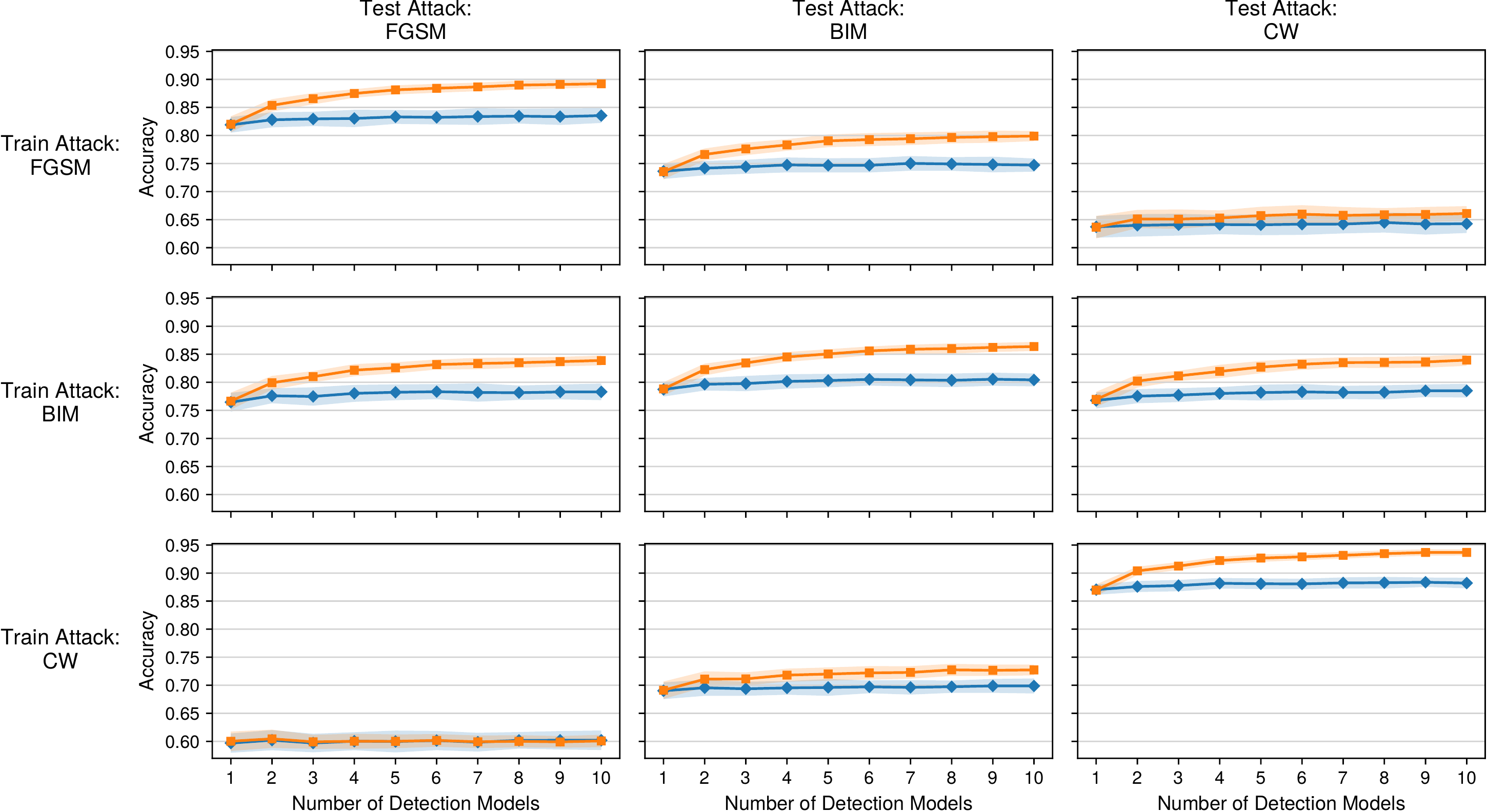}
  \rule{0pt}{4ex}  % Add additional spacing between image and legend
  {
    \fontsize{8}{10}  % The second arg, baseline-skip, should be approx 1.2x font size
    \fontfamily{phv}\selectfont  % phv is Helvetica (see helvet documentation)
    \begin{tabular}{cc}
      %\hskip2.6cm
      \diamond[0.1ex]{legend_blue} Control
      & \square[0.1ex]{legend_orange} Treatment
    \end{tabular}
  }
  \caption{
    Average model-wise adversarial input detection accuracies, where each point is calculated across
    100 trials. The sample standard deviations were added and subtracted from each sample mean to
    generate the shaded regions. The figure subplots each correspond to a specific attack used for
    the training data---as indicated by the leftmost labels---and a specific attack used for
    the test data---as indicated by the header labels. The endpoint values underlying the figure are
    provided in the appendix.
  }
  \label{fig:model_wise}
\end{figure*}

We use the 512-dimensional representation vectors extracted from the 1,024 representation models as
inputs to model-wise and unit-wise adversarial detectors---both treatment and control
configurations---as described in Section~\ref{sec:method}. All detection models are binary
classification neural networks that have a 100-dimensional hidden layer with a rectified linear unit
activation function. We did not tune hyperparameters, instead using the defaults as specified by the
library we employed, \texttt{scikit-learn}~\cite{scikit-learn}. Model-wise detectors differed in
their randomly initialized weights.

To evaluate the contribution of multiple models, we run experiments that vary 1)~the number of
detection models used for model-wise detection, and 2)~the number of units used for unit-wise
detection. For the treatment experiments, the number of underlying representation models matches
1)~the number of detection models for model-wise detection and 2)~the number of units for unit-wise
detection. For the control experiments, there is a single underlying representation model.

The number of units for the unit-wise control models was limited to 512, based on the dimensionality
of the penultimate layer representations. The number of units for the unit-wise treatment was
extended beyond this since its limit is based on the number of representation models, for which we
had more than 512. One way to incorporate more units into the unit-wise control experiments would be
to draw units from other network layers, but we have not explored that for this paper.

We are interested in the generalization capabilities of detectors trained with data from a specific
attack. While the training datasets we constructed were each limited to a single attack algorithm,
we separately tested each model using data attacked with each of the three algorithms---FGSM, BIM,
and CW.

For training and evaluating each detection model, the dataset consisted of 1)~the 9,125 images that
were originally correctly classified by the attacked model, and 2)~the 9,125 corresponding perturbed
variants. Models were trained with 90\% of the data and tested on the remaining 10\%. Each original
image and its paired adversarial counterpart were grouped, i.e., they were never separated such that
one would be used for training and the other for testing.

We retained all 9,125 perturbed images and handled them the same (i.e., they were given the same
class) for training and evaluation, including the instances that did not successfully deceive the
attacked model. For BIM and CW, the consequence of this approach is presumably minor, since there
were few unsuccessful attacks. For FGSM, which had a lower attack success rate, further work would
be needed to 1)~study the implications and/or 2)~implement an alternative approach.

We conducted 100 trials for each combination of settings. For each trial, random sampling was used
for 1)~splitting data into training and test groups, 2)~choosing representation models, and
3)~choosing which representation units to use for the unit-wise experiments.

\subsection{Hardware and Software}

The experiments were conducted on a desktop computer running Ubuntu 21.04 with Python 3.9. The
hardware includes an AMD Ryzen 9 3950X CPU, 64GB of memory, and an NVIDIA TITAN RTX GPU with 24GB of
memory. The GPU was used for training the CIFAR-10 classifiers and generating adversarial attacks.

The code for the experiments is available at~\ifsubmission{\url{https://anonymized/for/submission}}%
{\url{https://github.com/dstein64/multi-adv-detect}}.

\subsection{Results}

% ********************************************
% * <original>fig:model_wise 
% ********************************************

\paragraph{Model-Wise} Figure~\ref{fig:model_wise} shows average model-wise adversarial input
detection accuracies---calculated from 100 trials---plotted across the number of detection models.
The subplots represent different combinations of training data attacks and test data attacks. The
endpoint values underlying the figure are provided in the appendix.

\begin{figure*}[t]
  \centering
  \includegraphics[width=\linewidth]{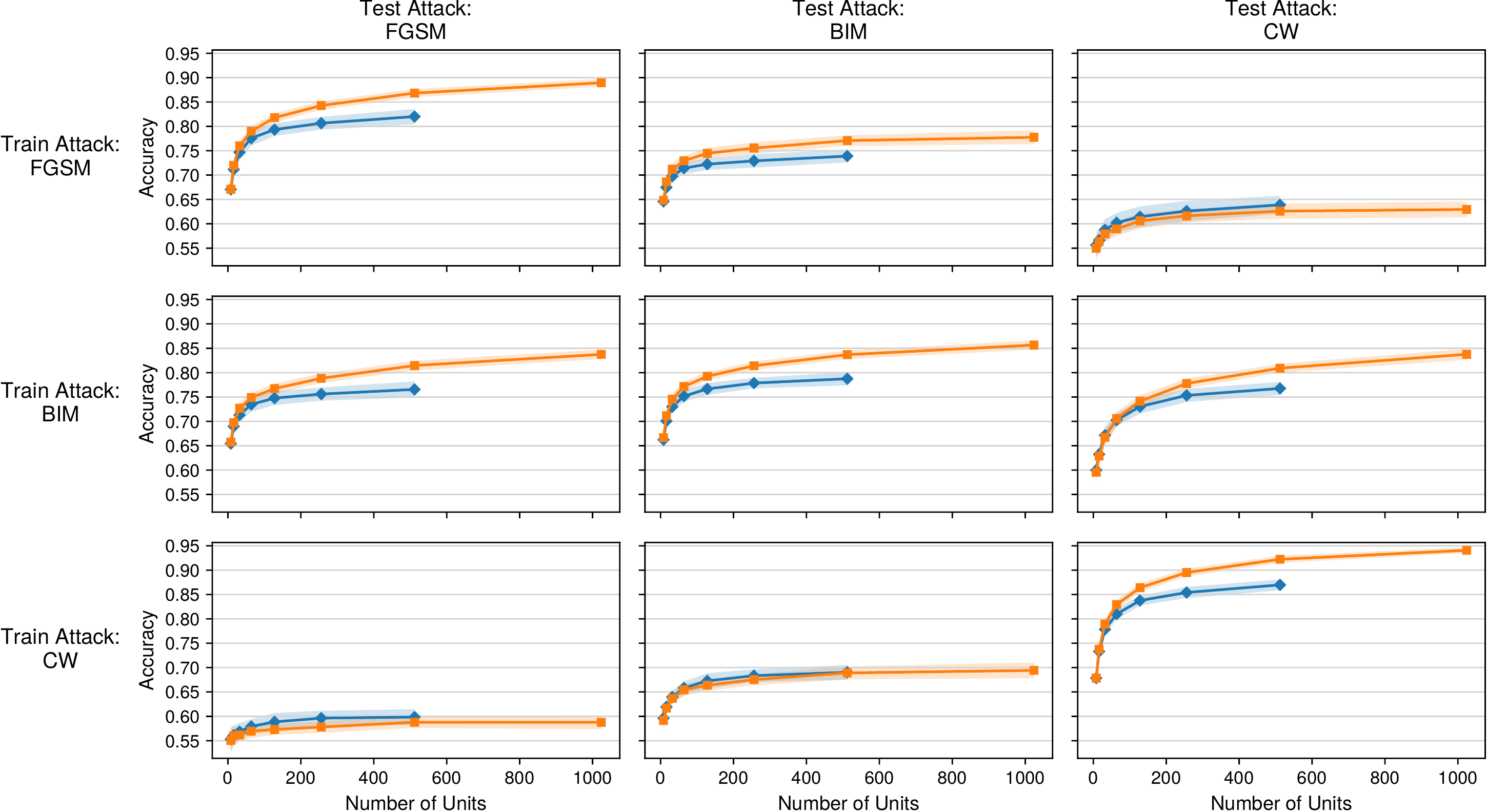}
  \rule{0pt}{4ex}  % Add additional spacing between image and legend
  {
    \fontsize{8}{10}  % The second arg, baseline-skip, should be approx 1.2x font size
    \fontfamily{phv}\selectfont  % phv is Helvetica (see helvet documentation)
    \begin{tabular}{cc}
      %\hskip2.6cm
      \diamond[0.1ex]{legend_blue} Control
      & \square[0.1ex]{legend_orange} Treatment
    \end{tabular}
  }
  \caption{
    Average unit-wise adversarial input detection accuracies, where each point is calculated across
    100 trials. The sample standard deviations were added and subtracted from each sample mean to
    generate the shaded regions. The figure subplots each correspond to a specific attack used for
    the training data---as indicated by the leftmost labels---and a specific attack used for
    the test data---as indicated by the header labels. The endpoint values underlying the figure are
    provided in the appendix.
  }
  \label{fig:unit_wise}
\end{figure*}

\paragraph{Unit-Wise} Figure~\ref{fig:unit_wise} shows average unit-wise adversarial input detection
accuracies---calculated from 100 trials---plotted across the number of units. The subplots represent
different combinations of training data attacks and test data attacks. The endpoint values
underlying the figure are provided in the appendix.

\section{Discussion}

Although subtle, for most scenarios the model-wise control experiments show an upward trend in
accuracy as a function of the number of detection models. This is presumably an ensembling effect
where there are benefits from combining multiple detection models even when they're each trained on
the same features. The model-wise treatment experiments tend to outpace the corresponding controls,
highlighting the benefit realized when the ensemble utilizes representations from distinct models.

The increasing accuracy for the unit-wise control experiments---as a function of the number of
units---is more discernible than for the corresponding model-wise control experiments (the latter
being a function of the number of models). The unit-wise gains are from having more units, and thus
more information, as discriminative features for detecting adversarial instances. In most scenarios
the treatment experiments---which draw units from distinct representation models---have higher
performance than the corresponding controls. An apparent additional benefit is being able to
incorporate more units when drawing from multiple models, not limited by the quantity of eligible
units in a single model. However, drawing units from multiple models also comes at a practical cost,
as it requires more computation relative to drawing from a single model.

As expected, detectors trained with data from a specific attack perform best when tested with data
from the same attack. Interestingly, detectors trained with BIM attack data appear to generalize
better relative to detectors trained with FGSM or CW attack data. This may be related to the
hyperparameters we used for each of the attacks, as opposed to being something representative of BIM
more generally.

\section{Related Work}
\label{sec:related_work}

We are aware of two general research areas that are related to what we've explored in this paper.
The approaches include 1)~the incorporation of ensembling for adversarial defense, and 2)~the usage
of hidden layer representations for detecting adversarial instances.

\subsection{Ensembling-Based Adversarial Defense}

Combining machine learning models is the hallmark of ensembling. For our work, we trained detection
models that process representations extracted from multiple independently trained models. For
model-wise detection, we averaged detection outputs across multiple models. Existing research has
explored ensembling techniques in the context of defending against adversarial
attacks~\cite{liu_deep_2019}. \citeauthor{bagnall_training_2017} train an ensemble---to be used for
the original task, classification, and also for adversarial detection---such that the underlying
models agree on clean samples and disagree on perturbed examples. The \emph{adaptive diversity
promoting regularizer}~\cite{pang_improving_2019} was developed to increase model diversity---and
decrease attack transferability---among the members of an ensemble. \citeauthor{abbasi_toward_2020}
devise a way to train ensemble \emph{specialists} and merge their predictions---to mitigate the risk
of adversarial examples.

% This table was moved here for better layout. See <original>table:model_wise for the original
% location in the LaTeX markup.
\begin{table*}[t]
  \caption{
    Average unit-wise adversarial input detection accuracies plus/minus sample standard deviations,
    calculated across 100 trials for each datum. These are a subset of values used to generate
    Figure~\ref{fig:model_wise}.
  }
  \label{table:model_wise}
  \addtolength{\tabcolsep}{-1.35pt}  % Reduce space between columns so table doesn't overflow
  \centering
  \begin{tabular}[b]{cccccccc}
    \toprule
    \multirow[b]{3}{*}[1.06pt]{\begin{tabular}{@{}c@{}} \\ Train \\ Attack\end{tabular}}
    & \multirow[b]{3}{*}[1.06pt]{\begin{tabular}{@{}c@{}} Number of \\ Detection \\ Models\end{tabular}}
    & \multicolumn{6}{c}{Test Attack} \\
    \cmidrule(r){3-8}
    & & \multicolumn{2}{c}{FGSM} & \multicolumn{2}{c}{BIM} & \multicolumn{2}{c}{CW} \\
    \cmidrule(r){3-4}
    \cmidrule(r){5-6}
    \cmidrule(r){7-8}
    &
    & \begin{tabular}{@{}c@{}}Control\end{tabular}
    & \begin{tabular}{@{}c@{}}Treatment\end{tabular}
    & \begin{tabular}{@{}c@{}}Control\end{tabular}
    & \begin{tabular}{@{}c@{}}Treatment\end{tabular}
    & \begin{tabular}{@{}c@{}}Control\end{tabular}
    & \begin{tabular}{@{}c@{}}Treatment\end{tabular} \\
    \midrule
    \multirow{2}{*}{FGSM}
    & 1 & \meansd{0.819}{0.014} & \meansd{0.820}{0.014}
    & \meansd{0.736}{0.014} & \meansd{0.735}{0.014}
    & \meansd{0.638}{0.019} & \meansd{0.637}{0.020} \\
    & 10 & \meansd{0.836}{0.013} & \meansd{0.892}{0.006}
    & \meansd{0.747}{0.012} & \meansd{0.799}{0.009}
    & \meansd{0.643}{0.017} & \meansd{0.661}{0.013} \\
    \addlinespace[1ex]
    \multirow{2}{*}{BIM}
    & 1 & \meansd{0.765}{0.017} & \meansd{0.766}{0.015}
    & \meansd{0.788}{0.013} & \meansd{0.788}{0.012}
    & \meansd{0.767}{0.014} & \meansd{0.770}{0.014} \\
    & 10 & \meansd{0.783}{0.015} & \meansd{0.839}{0.009}
    & \meansd{0.805}{0.012} & \meansd{0.864}{0.008}
    & \meansd{0.785}{0.012} & \meansd{0.840}{0.010} \\
    \addlinespace[1ex]
    \multirow{2}{*}{CW}
    & 1 & \meansd{0.597}{0.017} & \meansd{0.600}{0.017}
    & \meansd{0.690}{0.015} & \meansd{0.691}{0.016}
    & \meansd{0.870}{0.009} & \meansd{0.870}{0.010} \\
    & 10 & \meansd{0.602}{0.018} & \meansd{0.601}{0.011}
    & \meansd{0.699}{0.014} & \meansd{0.727}{0.010}
    & \meansd{0.883}{0.009} & \meansd{0.937}{0.005} \\
    \bottomrule
  \end{tabular}
\end{table*}

% This table was moved here for better layout. See <original>table:unit_wise for the original
% location in the LaTeX markup.
\begin{table*}[t]
  \caption{
    Average unit-wise adversarial input detection accuracies plus/minus sample standard deviations,
    calculated across 100 trials for each datum. These are a subset of values used to generate
    Figure~\ref{fig:unit_wise}.
  }
  \label{table:unit_wise}
  \centering
  \begin{tabular}[b]{cccccccc}
    \toprule
    \multirow[b]{3}{*}[1.06pt]{\begin{tabular}{@{}c@{}} \\ Train \\ Attack\end{tabular}}
    & \multirow[b]{3}{*}[1.06pt]{\begin{tabular}{@{}c@{}} \\ Number \\ of Units\end{tabular}}
    & \multicolumn{6}{c}{Test Attack} \\
    \cmidrule(r){3-8}
    & & \multicolumn{2}{c}{FGSM} & \multicolumn{2}{c}{BIM} & \multicolumn{2}{c}{CW} \\
    \cmidrule(r){3-4}
    \cmidrule(r){5-6}
    \cmidrule(r){7-8}
    &
    & \begin{tabular}{@{}c@{}}Control\end{tabular}
    & \begin{tabular}{@{}c@{}}Treatment\end{tabular}
    & \begin{tabular}{@{}c@{}}Control\end{tabular}
    & \begin{tabular}{@{}c@{}}Treatment\end{tabular}
    & \begin{tabular}{@{}c@{}}Control\end{tabular}
    & \begin{tabular}{@{}c@{}}Treatment\end{tabular} \\
    \midrule
    \multirow{3}{*}{FGSM}
    & 8 & \meansd{0.671}{0.014} & \meansd{0.671}{0.013}
    & \meansd{0.646}{0.012} & \meansd{0.648}{0.014}
    & \meansd{0.556}{0.024} & \meansd{0.550}{0.026} \\
    & 512 & \meansd{0.820}{0.016} & \meansd{0.868}{0.008}
    & \meansd{0.739}{0.013} & \meansd{0.771}{0.011}
    & \meansd{0.639}{0.019} & \meansd{0.626}{0.016} \\
    & 1,024 & -- & \meansd{0.890}{0.008}
    & -- & \meansd{0.778}{0.014}
    & -- & \meansd{0.629}{0.016} \\
    \addlinespace[1ex]
    \multirow{3}{*}{BIM}
    & 8 & \meansd{0.654}{0.013} & \meansd{0.657}{0.014}
    & \meansd{0.662}{0.012} & \meansd{0.667}{0.013}
    & \meansd{0.600}{0.019} & \meansd{0.596}{0.020} \\
    & 512 & \meansd{0.766}{0.017} & \meansd{0.815}{0.010}
    & \meansd{0.787}{0.014} & \meansd{0.837}{0.009}
    & \meansd{0.768}{0.013} & \meansd{0.809}{0.009} \\
    & 1,024 & -- & \meansd{0.838}{0.010}
    & -- & \meansd{0.857}{0.010}
    & -- & \meansd{0.838}{0.011} \\
    \addlinespace[1ex]
    \multirow{3}{*}{CW}
    & 8 & \meansd{0.553}{0.024} & \meansd{0.550}{0.026}
    & \meansd{0.596}{0.018} & \meansd{0.592}{0.019}
    & \meansd{0.679}{0.015} & \meansd{0.678}{0.017} \\
    & 512 & \meansd{0.599}{0.016} & \meansd{0.588}{0.012}
    & \meansd{0.690}{0.015} & \meansd{0.689}{0.013}
    & \meansd{0.870}{0.011} & \meansd{0.922}{0.007} \\
    & 1,024 & -- & \meansd{0.588}{0.014}
    & -- & \meansd{0.694}{0.016}
    & -- & \meansd{0.941}{0.006} \\
    \bottomrule
  \end{tabular}
\end{table*}

\subsection{Attack Detection from Representations}

For our research we've extracted representations from independently trained classifiers to be used
as features for adversarial example detectors. Hidden layer representations have been utilized in
various other work on adversarial instance detection. Neural network invariant
checking~\cite{ma_nic_2019} detects adversarial samples based on whether internal activations
conflict with invariants learned from non-adversarial data. \citeauthor{wojcik_adversarial_2020} use
hidden layer activations to train autoencoders whose own hidden layer activations---along with
reconstruction error---are used as features for attack detection. \citeauthor{li_adversarial_2017}
develop a cascade classifier that incrementally incorporates statistics calculated on convolutional
layer activations. At each stage, the instance is either classified as non-adversarial or passed
along to the next stage of the cascade that integrates features computed from an additional
convolutional layer. In addition to the methods summarized above, detection techniques have also
been developed that 1)~model the relative-positioned dynamics of representations passing through a
neural network~\cite{carrara_adversarial_2019}, 2)~use hidden layer activations as features for a
$k$-nearest neighbor classifier~\cite{carrara_detecting_2017}, and 3)~process the hidden layer units
that were determined to be relevant for the classes of interest~\cite{granda_can_2020}.

\section{Conclusion and Future Work}

We presented two approaches for adversarial instance detection---model-wise and unit-wise---that
incorporate the representations from multiple models. Using those two approaches, we devised
controlled experiments comprised of treatments and controls, for measuring the contribution of
multiple model representations in detecting adversarial instances. For many of the scenarios we
considered, experiments showed that detection performance increased with the number of underlying
models used for extracting representations.

The research leaves open various avenues for future work.

\begin{itemize}
  \item For our experiments, we trained 1,024 neural network representation models, whose diversity
  arises from using a different randomization seed for each. Perhaps other methods for imposing
  diversity would impact the performance of the detectors that depend on those models.
  \item It would be interesting to explore how existing adversarial defenses fare when extended to
  use multiple underlying models.
  \item Although we evaluated detectors across different attack algorithms, we always used data from
  a single attack for the purpose of training. Future research could investigate
  the effect of training with data from multiple attacks and/or varying hyperparameter settings for
  a specific attack.
  \item Our focus was on measuring the incremental gains of detecting attacks when incorporating
  multiple representation models. Further work could perform a thorough defense evaluation under
  more challenging threat models.
\end{itemize}

% **************************************
% * Appendix
% **************************************

\appendix

\section*{Appendix}

% Add Appendix link to Contents.
\addcontentsline{toc}{section}{Appendix}

The endpoint values underlying Figure~\ref{fig:model_wise} are included in
Table~\ref{table:model_wise}. The endpoint values underlying Figure~\ref{fig:unit_wise} are included
in Table~\ref{table:unit_wise}.

% ********************************************
% * <original>table:model_wise
% ********************************************

% ********************************************
% * <original>table:unit_wise
% ********************************************

% **************************************
% * References
% **************************************

{
% The AAAI formatting guidelines permit font size for references to be reduced to 9 point with 10
% point linespacing if the paper would otherwise exceed the number of allowed pages.
\fontsize{9}{10}\selectfont
\bibliography{paper}

\begin{thebibliography}{27}
\providecommand{\natexlab}[1]{#1}

\bibitem[{Abbasi et~al.(2020)Abbasi, Rajabi, Gagné, and
  Bobba}]{abbasi_toward_2020}
Abbasi, M.; Rajabi, A.; Gagné, C.; and Bobba, R.~B. 2020.
\newblock Toward {Adversarial} {Robustness} by {Diversity} in an {Ensemble} of
  {Specialized} {Deep} {Neural} {Networks}.
\newblock In Goutte, C.; and Zhu, X., eds., \emph{Advances in {Artificial}
  {Intelligence}}, Lecture {Notes} in {Computer} {Science}, 1--14. Cham:
  Springer International Publishing.
\newblock ISBN 978-3-030-47358-7.

\bibitem[{Akhtar and Mian(2018)}]{akhtar_threat_2018}
Akhtar, N.; and Mian, A. 2018.
\newblock Threat of {Adversarial} {Attacks} on {Deep} {Learning} in {Computer}
  {Vision}: {A} {Survey}.
\newblock \emph{IEEE Access}, 6: 14410--14430.
\newblock Conference Name: IEEE Access.

\bibitem[{Bagnall, Bunescu, and Stewart(2017)}]{bagnall_training_2017}
Bagnall, A.; Bunescu, R.; and Stewart, G. 2017.
\newblock Training {Ensembles} to {Detect} {Adversarial} {Examples}.
\newblock \emph{arXiv:1712.04006 [cs]}.

\bibitem[{Bengio, Lecun, and Hinton(2021)}]{bengio_deep_2021}
Bengio, Y.; Lecun, Y.; and Hinton, G. 2021.
\newblock Deep learning for {AI}.
\newblock \emph{Communications of the ACM}, 64(7): 58--65.

\bibitem[{Carlini and Wagner(2017)}]{carlini_towards_2017}
Carlini, N.; and Wagner, D. 2017.
\newblock Towards {Evaluating} the {Robustness} of {Neural} {Networks}.
\newblock \emph{arXiv:1608.04644 [cs]}.

\bibitem[{Carrara et~al.(2019)Carrara, Becarelli, Caldelli, Falchi, and
  Amato}]{carrara_adversarial_2019}
Carrara, F.; Becarelli, R.; Caldelli, R.; Falchi, F.; and Amato, G. 2019.
\newblock Adversarial {Examples} {Detection} in {Features} {Distance} {Spaces}.
\newblock In Leal-Taixé, L.; and Roth, S., eds., \emph{Computer {Vision} –
  {ECCV} 2018 {Workshops}}, Lecture {Notes} in {Computer} {Science}, 313--327.
  Cham: Springer International Publishing.
\newblock ISBN 978-3-030-11012-3.

\bibitem[{Carrara et~al.(2017)Carrara, Falchi, Caldelli, Amato, Fumarola, and
  Becarelli}]{carrara_detecting_2017}
Carrara, F.; Falchi, F.; Caldelli, R.; Amato, G.; Fumarola, R.; and Becarelli,
  R. 2017.
\newblock Detecting adversarial example attacks to deep neural networks.
\newblock In \emph{Proceedings of the 15th {International} {Workshop} on
  {Content}-{Based} {Multimedia} {Indexing}}, {CBMI} '17, 1--7. New York, NY,
  USA: Association for Computing Machinery.
\newblock ISBN 978-1-4503-5333-5.

\bibitem[{Dargan et~al.(2020)Dargan, Kumar, Ayyagari, and
  Kumar}]{dargan_survey_2020}
Dargan, S.; Kumar, M.; Ayyagari, M.~R.; and Kumar, G. 2020.
\newblock A {Survey} of {Deep} {Learning} and {Its} {Applications}: {A} {New}
  {Paradigm} to {Machine} {Learning}.
\newblock \emph{Archives of Computational Methods in Engineering}, 27(4):
  1071--1092.

\bibitem[{Goodfellow, Shlens, and Szegedy(2015)}]{goodfellow_explaining_2015}
Goodfellow, I.; Shlens, J.; and Szegedy, C. 2015.
\newblock Explaining and {Harnessing} {Adversarial} {Examples}.
\newblock In \emph{International {Conference} on {Learning} {Representations}}.

\bibitem[{Goodfellow et~al.(2014)Goodfellow, Pouget-Abadie, Mirza, Xu,
  Warde-Farley, Ozair, Courville, and Bengio}]{goodfellow_generative_2014}
Goodfellow, I.~J.; Pouget-Abadie, J.; Mirza, M.; Xu, B.; Warde-Farley, D.;
  Ozair, S.; Courville, A.; and Bengio, Y. 2014.
\newblock Generative {Adversarial} {Nets}.
\newblock In Ghahramani, Z.; Welling, M.; Cortes, C.; Lawrence, N.~D.; and
  Weinberger, K.~Q., eds., \emph{Advances in {Neural} {Information}
  {Processing} {Systems} 27}, 2672--2680. Curran Associates, Inc.

\bibitem[{Granda, Tuytelaars, and Oramas(2020)}]{granda_can_2020}
Granda, R.; Tuytelaars, T.; and Oramas, J. 2020.
\newblock Can the state of relevant neurons in a deep neural networks serve as
  indicators for detecting adversarial attacks?
\newblock \emph{arXiv:2010.15974 [cs]}.

\bibitem[{He et~al.(2016)He, Zhang, Ren, and Sun}]{he_deep_2016}
He, K.; Zhang, X.; Ren, S.; and Sun, J. 2016.
\newblock Deep {Residual} {Learning} for {Image} {Recognition}.
\newblock In \emph{2016 {IEEE} {Conference} on {Computer} {Vision} and
  {Pattern} {Recognition} ({CVPR})}, 770--778.

\bibitem[{Isola et~al.(2017)Isola, Zhu, Zhou, and
  Efros}]{isola_image--image_2017}
Isola, P.; Zhu, J.-Y.; Zhou, T.; and Efros, A.~A. 2017.
\newblock Image-{To}-{Image} {Translation} {With} {Conditional} {Adversarial}
  {Networks}.
\newblock In \emph{Proceedings of the {IEEE} {Conference} on {Computer}
  {Vision} and {Pattern} {Recognition} ({CVPR})}.

\bibitem[{Kingma and Ba(2014)}]{kingma_adam:_2014}
Kingma, D.; and Ba, J. 2014.
\newblock Adam: {A} {Method} for {Stochastic} {Optimization}.
\newblock \emph{arXiv:1412.6980 [cs]}.

\bibitem[{Krizhevsky(2009)}]{krizhevsky_learning_2009}
Krizhevsky, A. 2009.
\newblock Learning {Multiple} {Layers} of {Features} from {Tiny} {Images}.
\newblock Technical report.

\bibitem[{Krizhevsky, Sutskever, and Hinton(2012)}]{krizhevsky_imagenet_2012}
Krizhevsky, A.; Sutskever, I.; and Hinton, G.~E. 2012.
\newblock {ImageNet} {Classification} with {Deep} {Convolutional} {Neural}
  {Networks}.
\newblock In Pereira, F.; Burges, C. J.~C.; Bottou, L.; and Weinberger, K.~Q.,
  eds., \emph{Advances in {Neural} {Information} {Processing} {Systems} 25},
  1097--1105. Curran Associates, Inc.

\bibitem[{kuangliu(2017)}]{kuangliu_kuangliupytorch-cifar_2021}
kuangliu. 2017.
\newblock kuangliu/pytorch-cifar.
\newblock \url{https://github.com/kuangliu/pytorch-cifar}.

\bibitem[{Kurakin, Goodfellow, and Bengio(2017)}]{kurakin_adversarial_2017}
Kurakin, A.; Goodfellow, I.; and Bengio, S. 2017.
\newblock Adversarial examples in the physical world.
\newblock \emph{arXiv:1607.02533 [cs, stat]}.

\bibitem[{Li and Li(2017)}]{li_adversarial_2017}
Li, X.; and Li, F. 2017.
\newblock Adversarial {Examples} {Detection} in {Deep} {Networks} with
  {Convolutional} {Filter} {Statistics}.
\newblock In \emph{2017 {IEEE} {International} {Conference} on {Computer}
  {Vision} ({ICCV})}, 5775--5783.
\newblock ISSN: 2380-7504.

\bibitem[{Liu et~al.(2019)Liu, Wei, Chow, Loper, Gursoy, Truex, and
  Wu}]{liu_deep_2019}
Liu, L.; Wei, W.; Chow, K.-H.; Loper, M.; Gursoy, E.; Truex, S.; and Wu, Y.
  2019.
\newblock Deep {Neural} {Network} {Ensembles} {Against} {Deception}: {Ensemble}
  {Diversity}, {Accuracy} and {Robustness}.
\newblock In \emph{2019 {IEEE} 16th {International} {Conference} on {Mobile}
  {Ad} {Hoc} and {Sensor} {Systems} ({MASS})}, 274--282.
\newblock ISSN: 2155-6814.

\bibitem[{Ma et~al.(2019)Ma, Liu, Tao, Lee, and Zhang}]{ma_nic_2019}
Ma, S.; Liu, Y.; Tao, G.; Lee, W.-C.; and Zhang, X. 2019.
\newblock {NIC}: {Detecting} {Adversarial} {Samples} with {Neural} {Network}
  {Invariant} {Checking}.
\newblock In \emph{Proceedings 2019 {Network} and {Distributed} {System}
  {Security} {Symposium}}. San Diego, CA: Internet Society.
\newblock ISBN 978-1-891562-55-6.

\bibitem[{McCulloch and Pitts(1943)}]{mcculloch_logical_1943}
McCulloch, W.~S.; and Pitts, W. 1943.
\newblock A logical calculus of the ideas immanent in nervous activity.
\newblock \emph{The bulletin of mathematical biophysics}, 5(4): 115--133.

\bibitem[{Pang et~al.(2019)Pang, Xu, Du, Chen, and Zhu}]{pang_improving_2019}
Pang, T.; Xu, K.; Du, C.; Chen, N.; and Zhu, J. 2019.
\newblock Improving {Adversarial} {Robustness} via {Promoting} {Ensemble}
  {Diversity}.
\newblock In \emph{Proceedings of the 36th {International} {Conference} on
  {Machine} {Learning}}, 4970--4979. PMLR.
\newblock ISSN: 2640-3498.

\bibitem[{Papernot et~al.(2018)Papernot, Faghri, Carlini, Goodfellow, Feinman,
  Kurakin, Xie, Sharma, Brown, Roy, Matyasko, Behzadan, Hambardzumyan, Zhang,
  Juang, Li, Sheatsley, Garg, Uesato, Gierke, Dong, Berthelot, Hendricks,
  Rauber, and Long}]{papernot2018cleverhans}
Papernot, N.; Faghri, F.; Carlini, N.; Goodfellow, I.; Feinman, R.; Kurakin,
  A.; Xie, C.; Sharma, Y.; Brown, T.; Roy, A.; Matyasko, A.; Behzadan, V.;
  Hambardzumyan, K.; Zhang, Z.; Juang, Y.-L.; Li, Z.; Sheatsley, R.; Garg, A.;
  Uesato, J.; Gierke, W.; Dong, Y.; Berthelot, D.; Hendricks, P.; Rauber, J.;
  and Long, R. 2018.
\newblock Technical Report on the CleverHans v2.1.0 Adversarial Examples
  Library.
\newblock \emph{arXiv preprint arXiv:1610.00768}.

\bibitem[{Pedregosa et~al.(2011)Pedregosa, Varoquaux, Gramfort, Michel,
  Thirion, Grisel, Blondel, Prettenhofer, Weiss, Dubourg, Vanderplas, Passos,
  Cournapeau, Brucher, Perrot, and Duchesnay}]{scikit-learn}
Pedregosa, F.; Varoquaux, G.; Gramfort, A.; Michel, V.; Thirion, B.; Grisel,
  O.; Blondel, M.; Prettenhofer, P.; Weiss, R.; Dubourg, V.; Vanderplas, J.;
  Passos, A.; Cournapeau, D.; Brucher, M.; Perrot, M.; and Duchesnay, E. 2011.
\newblock Scikit-learn: Machine Learning in {P}ython.
\newblock \emph{Journal of Machine Learning Research}, 12: 2825--2830.

\bibitem[{Szegedy et~al.(2014)Szegedy, Zaremba, Sutskever, Bruna, Erhan,
  Goodfellow, and Fergus}]{szegedy_intriguing_2014}
Szegedy, C.; Zaremba, W.; Sutskever, I.; Bruna, J.; Erhan, D.; Goodfellow, I.;
  and Fergus, R. 2014.
\newblock Intriguing properties of neural networks.
\newblock In \emph{International {Conference} on {Learning} {Representations}}.

\bibitem[{Wójcik et~al.(2020)Wójcik, Morawiecki, Śmieja, Krzyżek, Spurek,
  and Tabor}]{wojcik_adversarial_2020}
Wójcik, B.; Morawiecki, P.; Śmieja, M.; Krzyżek, T.; Spurek, P.; and Tabor,
  J. 2020.
\newblock Adversarial {Examples} {Detection} and {Analysis} with {Layer}-wise
  {Autoencoders}.
\newblock \emph{arXiv:2006.10013 [cs, stat]}.

\end{thebibliography}
}

% Add References link to Contents. This has to be after \biliography, as the
% section is specified with that command.
\addcontentsline{toc}{section}{References}

\end{document}